\documentclass[runningheads]{llncs}

\usepackage{eccv}

\usepackage{eccvabbrv}

\usepackage{graphicx}
\usepackage{booktabs}
\usepackage{multirow}
\usepackage{colortbl}
\usepackage{bbm}
\usepackage{arydshln}
\usepackage{caption}
\usepackage{pifont}
\newcommand{\hollowstar}{\text{\ding{73}}}
\usepackage[accsupp]{axessibility}  

\usepackage{hyperref}

\usepackage{orcidlink}

\definecolor{cvprblue}{rgb}{0.21,0.49,0.74}
\definecolor{myYellow}{RGB}{255,239,213}
\definecolor{myBlue}{RGB}{231,239,252}
\definecolor{myOrange}{RGB}{255 127 14}
\definecolor{red}{RGB}{255,0,0}
\newcommand{\eccvVersion}[1]{{\color{black} #1}}
\begin{document}

\title{Rethinking Few-shot Class-incremental Learning: Learning from Yourself} 


\author{Yu-Ming Tang\inst{1,3}\orcidlink{0000-0001-5472-0079} \and
Yi-Xing Peng\inst{1,3}\orcidlink{0000-0003-4248-9820} \and
Jingke Meng\thanks{Corresponding Author}\inst{1,3}\orcidlink{0000-0001-5437-3070} \and
Wei-Shi Zheng\inst{1,2,3,4}\orcidlink{0000-0001-8327-0003} 
}

\authorrunning{Tang et al.}

\institute{School of Computer Science and Engineering, Sun Yat-sen University, China \and
Peng Cheng Laboratory, Shenzhen, China \and
Key Laboratory of Machine Intelligence and Advanced Computing, Ministry of Education, China \and
Guangdong Province Key Laboratory of Information Security Technology, Sun Yat-sen University, China \\
\email{\{tangym9, pengyx23\}@mail2.sysu.edu.cn, mengjke@gmail.com, wszheng@ieee.org}}

\maketitle
\begin{abstract}
Few-shot class-incremental learning (FSCIL) aims to learn sequential classes with limited samples in a few-shot fashion.
Inherited from the classical class-incremental learning setting, the popular benchmark of FSCIL uses averaged accuracy ($aAcc$) and last-task averaged accuracy ($lAcc$) as the evaluation metrics. 
However, we reveal that such evaluation metrics may not provide adequate emphasis on the novel class performance, and the continual learning ability of FSCIL methods could be ignored under this benchmark.
In this work, as a complement to existing metrics, we offer a new metric called generalized average accuracy ($gAcc$) which is designed to provide an extra equitable evaluation by incorporating different perspectives of the performance under the guidance of a parameter $\alpha$. 
We also present an overall metric in the form of the area under the curve (AUC) along the $\alpha$. 
Under the guidance of $gAcc$, we release the potential of intermediate features of the vision transformers to boost the novel-class performance.
Taking information from intermediate layers which are less class-specific and more generalizable, we manage to rectify the final features, leading to a more generalizable transformer-based FSCIL framework.
Without complex network designs or cumbersome training procedures, our method outperforms existing FSCIL methods at $aAcc$ and $gAcc$ on three datasets. See codes at\setcounter{footnote}{0}
\footnote{\url{https://github.com/iSEE-Laboratory/Revisting_FSCIL}}
\end{abstract}
\begin{figure}[t]
    \centering
      \includegraphics[width=0.9\linewidth]{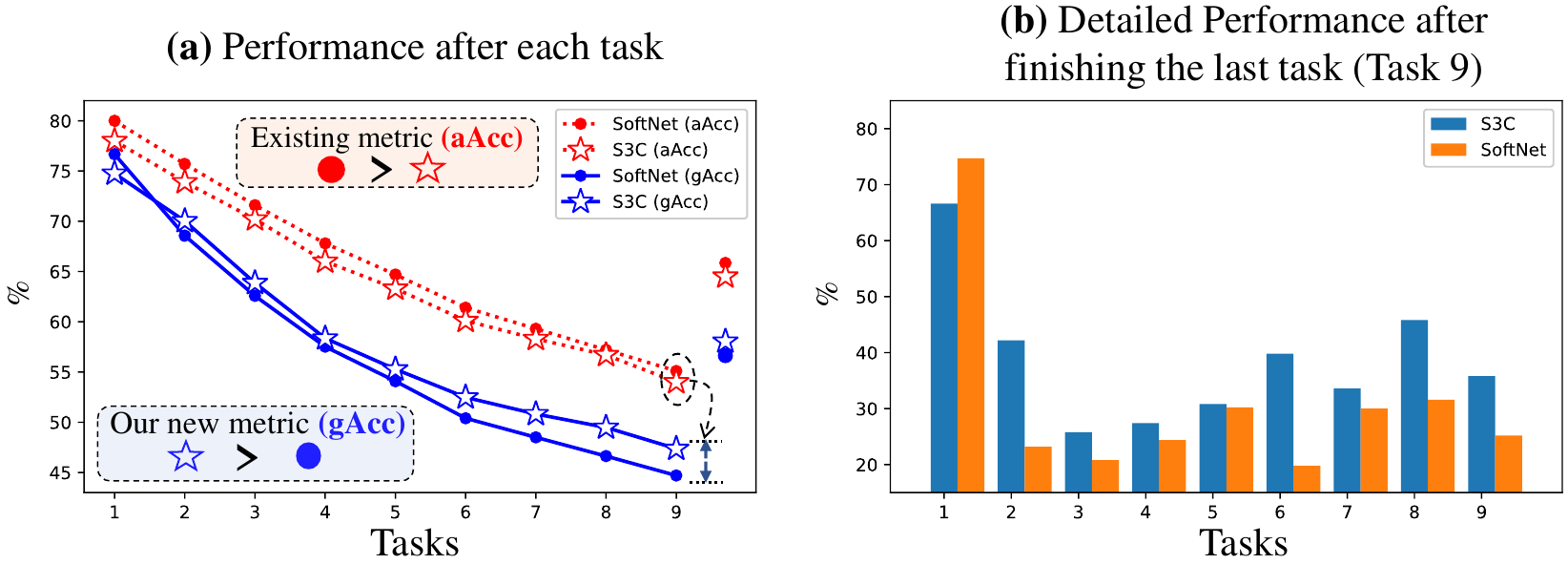}
       \captionsetup{font=scriptsize}
      \caption{
        Performance under different metrics of two recent methods~\cite{subnetwork, s3c}.
        \textbf{(a)}: The dotted line denotes the \textit{average accuracy (aAcc)} widely used in classical class-incremental learning. 
        The solid line presents our proposed \textit{generalized accuracy (gAcc)}.
        We also show the average of each point on the right side. (\textcolor{blue}{Blue}: $gAcc$, \textcolor{red}{Red}: $aAcc$). 
        \textbf{(b)}: The detailed accuracies of each task after training on the last task (task 8). 
        It is evident that S3C exhibits superior performance when confronted with novel classes. 
        The conventional metric $aAcc$ fails to reflect this due to the domination by the base-class performance.
        Models are trained on the CIFAR-100 dataset.  Best viewed in color. 
        \vspace{-20pt}
        }
    \label{fig:intro}
  \end{figure}
\vspace{-30pt}
\section{Introduction}
\vspace{-5pt}
\label{sec:intro}
Although existing deep neural networks (DNNs) achieve great success in various real-world applications\cite{lin2024rethinking, vaze2021open, radford2021learning, zheng2024versatile, li2024egoexo, li2024continual,fu2023asag, peng2022consistent}, they suffer from \textit{catastrophic forgetting} when dealing with sequential data streams.
The forgetting problem has attracted heavy attention in recent years~\cite{apg, imagine, PASS, SDC, SSRE, podnet, inthewild, dmc, icarl, lwf, zhao2020maintaining, ucir, castro2018end, der, smith2021always}.
In this work, we focus on a more realistic and challenging problem: few-shot class-incremental learning (FSCIL), in which after training on a task with sufficient data (\textbf{base} task), only a few samples are provided during incremental tasks (\textbf{novel} tasks).
Since Tao \etal ~\cite{topic} proposed the FSCIL benchmark in 2020, a large number of algorithms have emerged to tackle this problem, 
including meta-learning based~\cite{metafscil, cfscil, limit},
structure based~\cite{subnetwork, ahmad2022few, topic, cec, dsn}, 
feature space based~\cite{regularizer, cfscil, neural_collapse, fact, s3c, kim2022warping,chen2020incremental} methods, \etc.

However, almost all these methods are evaluated following the traditional class-incremental learning setting, \ie using the average accuracy ($aAcc$) or the last-task averaged accuracy ($lAcc$) as the main metrics to critique the proposed method.
Such accuracies are calculated by weighted averaging the base-task and novel-tasks accuracies with the weight proportional to the number of classes of each task.
Considering the reality that the FSCIL benchmarks make the majority of the classes (about 50\% to 60\%) as the base task to establish a well-trained backbone for few-shot continual learning, the existing $aAcc$ or $lAcc$ metric is \textbf{dominated} by the base-class performance.
In other words, current evaluation metrics can be largely boosted by only improving the performance of base classes.
As shown in \cref{fig:intro}, a method like softNet~\cite{subnetwork} with higher base-class performance (see \cref{fig:intro} (\textcolor{red}{b})) usually gets higher 
$aAcc$ (see red lines in \cref{fig:intro} (\textcolor{red}{a})) regardless of the inferior performance on the novel classes.
Due to the inadequacy to capture performance changes in novel classes, current metrics are biased toward the performance of base classes and have limitations in guiding the FSCIL community.
To tackle this, we propose a more balanced metric called \textit{generalized accuracy (gAcc)}.
Parameterized by $\alpha$ ranging from 0 to 1, the $gAcc$ offers a way to evaluate a model with explicit emphasis on performance on novel classes.
Furthermore, we can comprehensively evaluate the model by calculating the area under the curve (AUC) along the parameter $\alpha$.
\eccvVersion{It is worth mentioning that our intention is NOT to replace all previous metrics,
but to provide a more fine-grained and balanced evaluation of FSCIL methods complementary to them.}

We empirically find that FSCIL can benefit from the powerful vision transformer (ViT)~\cite{vit} by analyzing the performance of novel classes.
Specifically, although vanilla ViT performs poorly in terms of unseen novel classes, we reveal that the intermediate features of ViT are \textbf{less} specified on base classes. 
It means the ViT can learn abundant semantic information but the information is mistakenly dropped in deep layers when extracting features for novel classes.

Based on the observation, we propose a lightweight Feature Rectification (FR) module to rectify the output features of the ViT conditioned on the intermediate features.
To achieve this, the FR module is supervised with two losses that work on two different levels, namely instance-level relation transfer and class-center relation transfer.
The former transfers the relations between instances while the latter focuses on relations between instances and the class prototypes.
We further enhance the FR module by leveraging hierarchical knowledge from multiple intermediate layers.
Finally, our method boosts the novel-class accuracy and maintains a comparable performance of base classes.

Our contributions are summarized as follows.
(1) We revisit the existing FSCIL benchmark and propose a novel metric called \textit{generalized accuracy (gAcc)} for a more fine-grained and balanced evaluation.
We have reimplemented 14 FSCIL methods with codes released and show that the newly introduced SOTAs do not necessarily perform better on $gAcc$. 
(2) 
Based on $gAcc$, we reveal the potential of vision transformers in FSCIL.
We empirically find that although features from the intermediate layer exhibit sub-optimal performance for the base classes, their transferrable ability helps them to have a distinct advantage in handling novel classes.
(3) A lightweight FR module is proposed to rectify the final features to be more transferrable.
The FR module manages to make the output feature more generalizable with assistance from intermediate features.
(4) We conduct extensive experiments on 3 datasets and evaluate our method under both $aAcc$ and $gAcc$, showing the effectiveness and superiority of our method.
\vspace{-10pt}
\section{Related Work}
\vspace{-10pt}
\label{sec:related}
\noindent\textbf{Meta-learning based methods}.
Due to the limited samples of novel classes, some works~\cite{metafscil, cfscil, limit, zheng2021few} propose to use meta-learning to improve the transferability of the model.
Concretely, MetaFscil~\cite{metafscil} proposed to simulate novel tasks by sampling from the base task.
C-FSCIL~\cite{cfscil} uses meta-learning to train the backbone, which maps images from different classes to quasi-orthogonal vectors.
ALFSCIL~\cite{li2024analogical} formulate the process into a bi-level optimization problem.

\noindent\textbf{Optimization-based methods}.
There is a large group of works focusing on the optimization of the DNNs in FSCIL.
CLOM~\cite{clom} considers the behavior of the backbone under different margins of the cosine loss.
NC-FSCIL~\cite{neural_collapse} designs a pre-defined classifier for efficient FSCIL without any explicit regularizer.
SPPR~\cite{sppr} proposes a prototype refinement strategy for stronger expression ability.
FACT~\cite{fact} assigns virtual prototypes in the current stage for future updates.
Alice~\cite{alice} instead focuses on the compactness and the variety of the feature space.
Based on the classifier learning, S3C~\cite{s3c} proposes a self-supervised stochastic classifier to tackle the overfitting and forgetting problem.
There are more works working on the feature space constraint~\cite{regularizer, kim2022warping,chen2020incremental,yao2022few, wang2024few, Deng_2024_WACV, kim2024mics}, or 
knowledge distillation~\cite{cui2022uncertainty, cheraghian2021semantic, dong2021few, BiDistFSCIL, zhu2022feature, cui2021semi,pan2023ssfe, cui2023uncertainty}.

\noindent\textbf{Data-driven methods.}
There is also a branch of work focusing on the data side of this problem, which focuses on the data replay~\cite{data_free, agarwal2022semantics}.
FSIL-GAN~\cite{agarwal2022semantics} employs a GAN to generate features to compensate for the scarcity of data.
DF-replay~\cite{data_free} uses synthesized data generated by a generator to replay.
PL-FSCIL~\cite{jiang2024few} seeks prior knowledge from unlabeled data.

\noindent\textbf{Structure-based methods}.
In addition to these works, a unique branch of works focuses on the network structure, like topology preserving~\cite{topic}, growing classifier~\cite{cec}, proposing a dynamic structure of the backbone~\cite{dsn}, structure fusion~\cite{ahmad2022few, 10018890}, sub-networks\cite{subnetwork} and ensemble models\cite{zhu2024enhanced}.

Compared with existing works, our work rephrases the challenge of the FSCIL setting and first reveals the potential of the intermediate features of the ViT backbone.
There is also an existing work~\cite{alice} offering an alternative metric. They did not provide a fine-grained evaluation with various perspectives and have potential risks in stability and symmetry problems.
Please find more detailed discussions about related works in the supplementary material.
\vspace{-13pt}
\section{Generalized Average Accuracy}
\vspace{-10pt}
\label{sec:gacc}
\subsection{Limitation of the Existing Metrics}

\noindent\textbf{Accuracy metric on incremental tasks.} 
Consider a classification model $f(\cdot)$ that takes an image as input and outputs the class predictions.
For an incremental problem, there are multiple tasks and multiple test sets. 
Let us define test set at task $\mathcal{T}_i$ as $\mathcal{D}_{test}^i = \{x_t^i, y_t^i\}_{t=1}^{n^i}$. 
Usually, the labeling space $\mathcal{Y}^i \triangleq  \{y_t^i | (x_t^i, y_t^i) \in \mathcal{D}_{test}^i\}$ of every task does not have any intersection, \ie $\forall i \neq j, \mathcal{Y}^i\bigcap\mathcal{Y}^j = \emptyset$.
After trained on $\mathcal{T}_i$, the model can be denoted as $f_i(\cdot)$. 
The corresponding accuracy of model $f_i(\cdot)$ at task $\mathcal{T}_j$ ($j\leq i$) can be denoted as $\mathcal{A}^j_i \triangleq \frac{1}{n^j}\sum_{t=1}^{n^j} \mathbbm{1}(\zeta(f_i(x_t^j)) = y_t^j)$, where $\mathbbm{1}(\cdot) = 1$ if the condition is true, otherwise 0. $\zeta$ is the argmax operation.

\noindent\textbf{Average accuracy.} 
It is worth mentioning that for class-incremental learning, evaluating the model on \textbf{all} the previous tasks is essential.
To test the model on previous tasks, an overall test set that contains all previous test sets is constructed as $\mathcal{D}_{test}^{1\rightarrow i} = \bigcup_{m=1}^{i} \mathcal{D}_{test}^m$. 
After trained on task $\mathcal{T}_i$, the accuracy of model $f_i(\cdot)$ on the overall test set $\mathcal{D}_{test}^{1\rightarrow i}$ can be presented as $A_i^{1 \rightarrow i}$.
Once a $n_t$-task incremental training ends, the \textit{average accuracy (aAcc)} can be calculated as:
\begin{equation}
    aAcc = \frac{1}{n_t}  \sum_{i=1}^{n_t} aAcc_i =\frac{1}{n_t} \sum_{i=1}^{n_t} A_i^{1\rightarrow i},
\end{equation}
where the $aAcc_i$ is the average accuracy at each task.
We can also derive two additional metrics incidentally: last-task averaged accuracy $lAcc = aAcc_{n_t} = A_{n_t}^{1\rightarrow n_t}$,
and task-wise average accuracy ($tAcc$):
\begin{equation}
    tAcc = \frac{1}{n_t} \sum_{i=1}^{n_t} tAcc_i = \frac{1}{n_t}  \sum_{i=1}^{n_t} \frac{1}{i} \sum_{j=1}^{i} A^j_i,
\end{equation}
where the $tAcc_i$ is the task-wise accuracy at each task.
Note that the $aAcc$ does not share the same value as $tAcc$ normally since the number of classes of tasks is usually not the same.
This will be elaborated on later in detail.

\noindent\textbf{Limitation.}
Since FSCIL usually sets the majority of the classes as the base task, existing metrics potentially neglect the performance changes of novel classes.
To elaborate on this, we analyze the $aAcc$ as follows.
Firstly, we denote the size of the labeling space of task $\mathcal{T}_i$ as $|\mathcal{Y}_i|$.
In FSCIL, the number of classes of each novel task is usually the same, \ie $|\mathcal{Y}_{novel}| = |\mathcal{Y}_2| = |\mathcal{Y}_3| =\cdots =|\mathcal{Y}_{n_t}| \ll  |\mathcal{Y}_1|$.
Assuming there are total $n_t$ tasks, and task $\mathcal{T}_1$ is the \textit{base task} with sufficient classes and samples (non-few-shot).
The following tasks $\{\mathcal{T}_{i}\}_{i=2}^{n_t}$ are \textit{novel tasks}, with $|\mathcal{Y}_{novel}|$ classes for each task and a few samples per class.
Now, we can rewrite average accuracy on task $\mathcal{T}_i$ as:
\begin{equation}
    aAcc_i    = \frac{|\mathcal{Y}_1| A^1_i  + |\mathcal{Y}_{novel}| \sum_{j=2}^i  A^j_i}{|\mathcal{Y}_1| + (i-1) |\mathcal{Y}_{novel}|}
              =\frac{\frac{|\mathcal{Y}_1|}{|\mathcal{Y}_{novel}|} A^1_i  +  \sum_{j=2}^i  A^j_i}{\frac{|\mathcal{Y}_1|}{|\mathcal{Y}_{novel}|} + (i-1) }.
    \label{eq:acci}
\end{equation}
From \cref{eq:acci} we see that the value of $aAcc_i$ will be dominated by the base class accuracy $A_i^1$ if the ratio $\frac{|\mathcal{Y}_1|}{|\mathcal{Y}_{novel}|}$ is large.
Unfortunately, the ratio is indeed large in the FSCIL settings. (\eg $\frac{|\mathcal{Y}_1|}{|\mathcal{Y}_{novel}|}=12$ for both CIFAR-100 and \emph{mini}ImageNet dataset, 10 for CUB-200.)
Under such a large ratio, the changes (both improvement and degrades) of novel classes cannot be effectively evaluated by $aAcc$.

\vspace{-10pt}
\subsection{Generalizing the Average Accuracy}
\vspace{-3pt}
To overcome the limitations of existing metrics, we now define the \textit{generalized accuracy} ($gAcc$) at task $\mathcal{T}_i$ as:
\begin{equation}
    gAcc_i(\alpha) = \frac{\alpha\frac{|\mathcal{Y}_1|}{|\mathcal{Y}_{novel}|} A^1_i  +  \sum_{j=2}^i  A^j_i}{\alpha\frac{|\mathcal{Y}_1|}{|\mathcal{Y}_{novel}|} + (i-1) },
    \label{gacc_alpha}
\end{equation}
in which the $gAcc$ is parameterized by $\alpha \in [0,1]$.
At task $\mathcal{T}_i$, $gAcc_i(0)$ measures the accuracy of the novel tasks only.
$gAcc_i(\frac{|\mathcal{Y}_{novel}|}{|\mathcal{Y}_1|})$ denotes the task-wise accuracy $tAcc_i$ at task $\mathcal{T}_i$ while $gAcc(1)$ represents the average accuracy $aAcc_i$ at task $\mathcal{T}_i$.
In addition to these discrete values, we further generalize the $\alpha$ to any rational numbers in $[0,1]$. 
\eccvVersion{Controlled by $\alpha$ in \cref{gacc_alpha}, when $\alpha \rightarrow 0$, $gAcc(\alpha)$ emphasizes to novel class while $\alpha \rightarrow 1$, $gAcc(\alpha)$ emphasizes the base class performance, \ie ,}
the larger the $\alpha$, the greater the influence of the base class on the value of $gAcc(\alpha)$.
To evaluate the performance of a model on a specific task $\mathcal{T}_i$, a standalone value is needed.
Thus, we introduce the area under the curve (AUC) of $gAcc_i(\alpha)$ along the parameter $\alpha$ as a comprehensive metric:  
\begin{equation}
    gAcc_i = \int_{0}^{1} gAcc_i(\alpha) \, d\alpha.
    \label{eq:gacci}
\end{equation}
Similar to $aAcc$, for multiple tasks, we average the $gAcc_i$ for each task and get an overall metric $gAcc$ for the whole learning process:
\begin{equation}
\small
    gAcc = \frac{1}{n_t}  \sum_i^{n_t} gAcc_i.
    \label{eq:gacc}
\end{equation}
The proposed $gAcc$ provides different evaluation perspectives using different $\alpha$ so we believe such a metric can provide a more balanced evaluation of FSCIL.
\begin{figure}[t]
    \centering
      \includegraphics[width=0.85\linewidth]{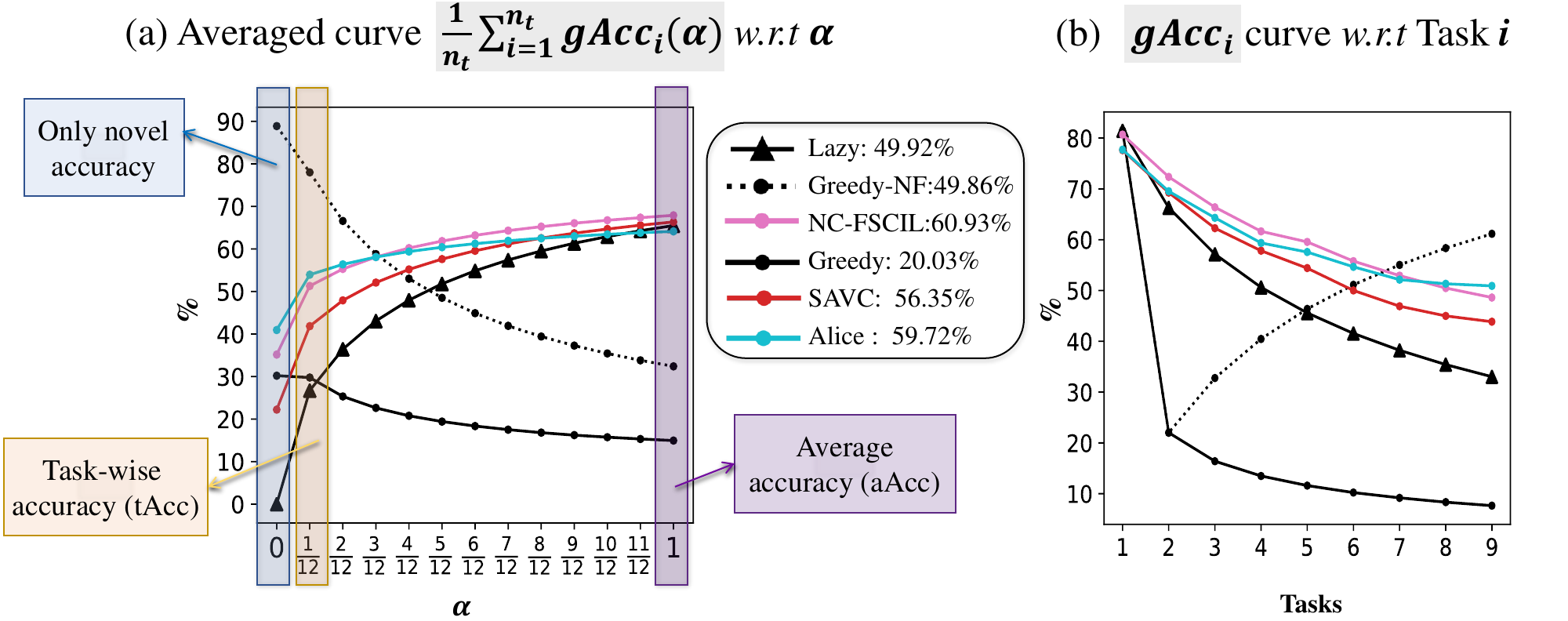}
      \captionsetup{font=scriptsize}
      \caption{
      FSCIL performances are shown in \textit{generalized accuracy}.
      (a): The $gAcc$ curve (averaged across all $n_t$ tasks) v.s. param $\alpha$.
      (b): The $gAcc$ AUC of each task $\mathcal{T}_i$ (\cref{eq:gacci}).
      In the legend, we show the average AUC across tasks(\cref{eq:gacc}) of each method.
      We evaluate recent works including SAVC\cite{savc}, NC\cite{neural_collapse}, and Alice\cite{alice} on the \emph{mini}ImageNet dataset.
      Some corner cases: 
      \textbf{Lazy}: The model maintains base-task performance while refusing to learn anything from novel tasks. 
      \textbf{Greedy}: Regardless of the previous knowledge, the model only greedily focuses on the current task. 
      \textbf{Greedy-NF}: A \textbf{N}on-\textbf{F}orget version of `Greedy'.
      See more about these cases in our supplementary material.
      }
    \label{fig:gacc}
    \vspace{-15pt}
  \end{figure}

\vspace{-10pt}
\subsection{Analysis}
We further study the necessity of the proposed $gAcc$.
We evaluate recent SOTAs in FSCIL including SAVC~\cite{savc}, Alice~\cite{alice} and NC-FSCIL~\cite{neural_collapse}, and the results are in \cref{fig:gacc}.
We show that the $gAcc$ provides more fine-grained evaluation than the classical accuracy metric.
For example, in the left part of \cref{fig:gacc}, although the $gAcc(1)$, which is equal to the $aAcc$, of SAVC is close to that of NC-FSCIL, 
the area under the curve(AUC) of SAVC is much smaller than NC-FSCIL.
This indicates that the novel class performance is much worse for SAVC, which cannot be discovered by conventional metrics.

Furthermore, to show the superiority of our metric, we consider three corner cases: Lazy, Greedy, and Greedy-NF.
`Lazy' stands for a type of algorithm that \textbf{only} focuses on maintaining \textbf{base-class} performance, \ie $\forall i \geq j>1, A^j_i=0, A_i^1 = A_1^1$.
By setting $A_1^1=85\%$, we can see that when $\alpha = 1$ (\ie the traditional $aAcc$ metric), its performance is close to SOTA methods.
\eccvVersion{Further, when considering popular forgetting metrics like performance dropping ($PD = A_i^j - A_j^j$) and Knowledge rate ($KR = A_i^j / A_j^j$), because the forgetting is zero, its PD is 0 and KR is 1.
All these metrics indicate such `Lazy' model is superior, which is contradictory to common sense.
However, our proposed $gAcc$ reveals that its performance is far from satisfying by setting $\alpha < 1$ and its AUC is far smaller than other methods.}
In other words, using the traditional metrics will indicate algorithms like `Lazy' achieving SOTA performance while such methods perform poorly evaluated by our proposed $gAcc$.
The $gAcc$ can also discover other corner cases.
`Greedy' is to greedily learn the current knowledge (\ie $A^1_1 = 85\%$, $A_i^j = 100\%$, when $i = j$ else $0$).
``Greedy'' achieves low both $aAcc$ and $gAcc$.
We further eliminate the forgetting problem from `Greedy' and define `Greedy-NF' on top of it: 
set accuracy $A_i^j$ to $100\%$ for $\forall j > 1,  j<i $ (no forgetting at novel classes).
We find that although `Greedy-NF' performs well when majorly considering the novel-class performance ($\alpha \in \{0,\frac{1}{12}\,\frac{2}{12}\,\frac{3}{12}\}$), the overall performance (AUC) is not satisfying.
Please refer to Sec. \textcolor{red}{A1} in the supplementary material for more details about these corner cases.

By analyzing these cases, we show the superior properties of the proposed $gAcc$. 
Firstly, the $gAcc$ metric effectively distinguishes methods (\eg `Lazy') that only retain the base-task performance, thereby encouraging methods to learn from novel classes.
Besides, $gAcc$ avoids overemphasis on the performance of base classes or novel classes (corner case `Greedy-NF' is still far worse than normal methods in terms of AUC), providing a more balanced evaluation outcome.
In other words, the proposed $gAcc$ metric provides objective and balanced results when evaluating FSCIL methods, eliminating `shortcuts' in $aAcc$ that focus only on the performance of the base classes. 

In conclusion, the $gAcc$ curve and the AUC are necessary and powerful tools to evaluate the FSCIL methods.

\begin{figure}[t]
    \centering
      \includegraphics[width=0.75\linewidth]{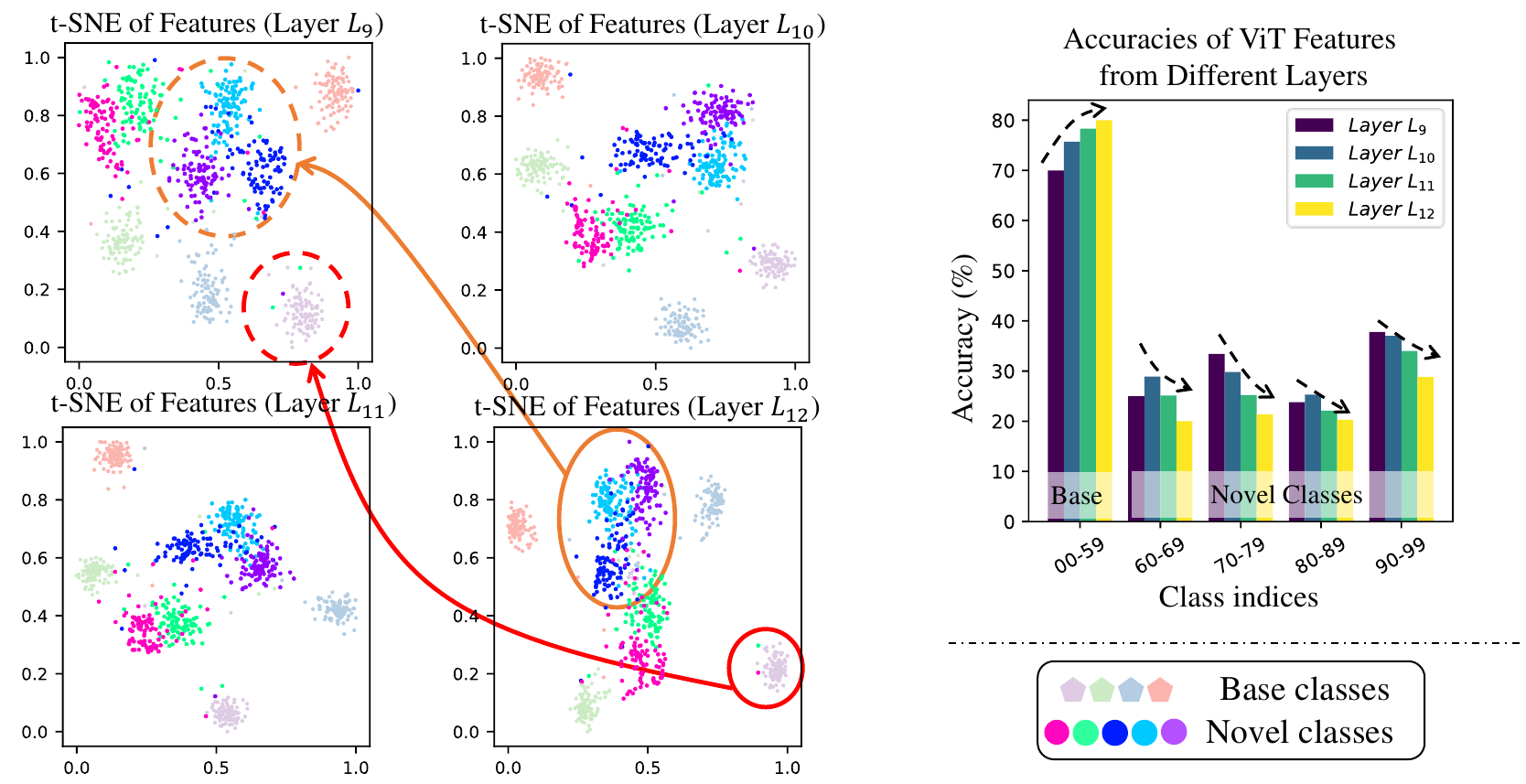}
       \captionsetup{font=scriptsize}
      \caption{
      t-SNE~\cite{tsne} visualization of \emph{mini}ImageNet test set features from different layers of a ViT.
      \textbf{Left:} we choose layers $L_9$, $L_{10}$, $L_{11}$, $L_{12}$(final) and show both base classes and novel classes features.
      It is observed that shallow features are more dispersed than deep features.
      \textbf{Right:} Accuracies of different tasks on features from various layers.
      It is clear that intermediate features achieve better performance on novel classes and thus have better generalization ability.
      Best viewed in color.
      \vspace{-15pt}
      }
    \label{fig:tsne_method}
  \end{figure}
\vspace{-10pt}
\section{Learn from Yourself}
\vspace{-5pt}
\subsection{Preliminary and The FSCIL Setting}
\label{sec:idea}
For task $\mathcal{T}_i \in \{\mathcal{T}_i\}_{i=1}^{T}$, the training set can be denoted as $\mathcal{D}_{train}^i = \{x_t^i, y_t^i\}_{t=1}^{m^i}$.
For FSCIL settings, the base task $\mathcal{T}_1$ contains $|\mathcal{Y}_1|$ classes and sufficient samples for training.
For $\forall i > 1$, there are $|\mathcal{Y}_{novel}|$ classes for each task and $k$ samples per class. 
Typically, $|\mathcal{Y}_{novel}|$ is set to $5$, and $k=5$, \ie a 5-way 5-shot setting.
We rephrase the classification model $f(\cdot)$ in \cref{sec:gacc} as $f(\cdot) = \Phi(f_B(\cdot))$ where $\Phi(\cdot)$ is the classifier and $f_B(\cdot)$ is the backbone.
We follow previous works to use a cosine classifier: $\Phi(X) = P^\top \overline{X}$, where $P\in \mathbb{R}^{d\times |\mathcal{Y}|}$ and $\overline{X} \in \mathbb{R}^{d\times 1}$ are the L2-normalized inputs, and $d$ is the feature dimension of  $f_B(\cdot)$, $|\mathcal{Y}| = \sum_i |\mathcal{Y}_i|$ is the number of classes seen so far.
In this work, we employ the Vision Transformer(ViT)~\cite{vit} in the FSCIL setting.
For a ViT with $N_L$ layers $\{L_l\}_{l=1}^{N_L}$, given a input image $x$, the output can be denoted as $X_{N_L} = L_{N_L}(L_{N_L-1}(\cdots L_1(x))$.

\noindent\textbf{Baseline.} Recent works~\cite{cec, f2m, savc, neural_collapse, s3c, mazumder2021few, clom, cfscil, regularizer} suggest a training-frozen scheme, which trains the backbone $f_B(\cdot)$ and the classifier $\Phi(\cdot)$ (if available) on the base-task only, and then freeze the backbone during novel tasks. 
For novel classes, the classifier is extended by mean features (\ie prototypes) extracted by the frozen backbone.
We follow and employ the scheme as our baseline.

\begin{figure*}[t]
    \centering
      \includegraphics[width=\linewidth]{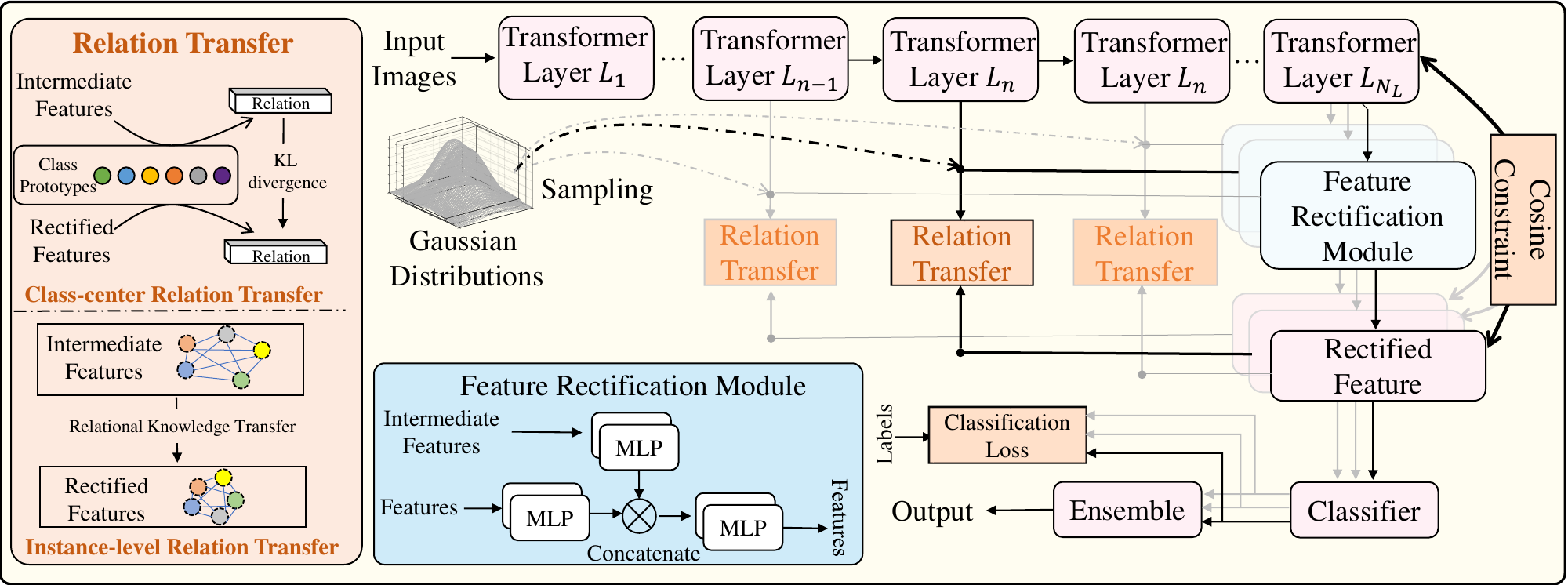}
       \captionsetup{font=scriptsize}
      \caption{
      The overall framework of our method.
      The Feature Rectification (FR) module takes the final and the intermediate features as input and outputs the rectified features for classification.
      The relation transfer losses are designed to convey valuable information from intermediate features to the rectified ones.
      The details of these two parts are shown in the left and lower parts of the figure.
      Further, a cosine constraint is proposed to maintain the base class performance and a classification loss is employed to adapt to the novel classes.
      Best viewed in color.
      \vspace{-15pt}
      }
    \label{fig:method}
  \end{figure*}
\noindent\textbf{Main idea of our method.}
Given that the backbone is initially trained for the base task and remains fixed during novel tasks, the main challenges lie in the potential overfitting on the base task and the poor feature extraction ability of the novel classes.
We conduct a toy experiment to show these problems and show that the intermediate feature can convey valuable information to tackle these problems.
Results are shown in \cref{fig:tsne_method}, where we select 4 base classes and 5 novel classes and utilize the t-SNE~\cite{tsne} to visualize the features from different layers.
It can be concluded that features extracted by the base-classes trained backbone are overfitted to the base classes and cannot be generalized well on novel classes, \ie the final features are well aggregated, but it is hard to distinguish among unseen classes. (see $L_{12}$ features on the left and the corresponding performance on the right).
Compared with the final features (\ie$L_{12}$), the shallower intermediate features ($L_9$, $L_{10}$, $L_{11}$) are more dispersed, which makes them more discriminative for unseen classes.
This is cross-verified by the performance shown on the right side of \cref{fig:tsne_method}.
We can see that using intermediate features can largely boost the novel classes' performance (up to 15\% improvement) with a price of performance decline of the base classes.
Based on this observation, we propose to rectify the final feature based on the intermediate feature to better handle both base and novel classes.

\subsection{Feature Rectification}
\label{sec:method}
We propose a Feature Rectification (FR) module to rectify the final feature based on the relation information from the intermediate features.
Specifically, given a pair of final feature $X_{N_L}$ and the intermediate feature $X_l$, we design an FR module conditioned on the intermediate features:
\begin{equation}
    X_{FR} = M_{mix}(\text{cat}(M_{F}(X_{N_L}), M_{I}(X_l)),
\end{equation}
where `cat' means the concatenate operation. $M_{mix}$, $M_F$ and $M_I$ are 2-layer MLPs.
By introducing the intermediate features, the FR module has access to a reference for extracting the valuable structural information lies in the intermediate features.
A naive way to mine such information is to directly pull rectified features closer to shallow features on an instance-by-instance basis.
\eccvVersion{Although such an instance-wise loss can bridge shallow features and rectified features, directly pulling the rectified feature to the shallow one could lead to noisy features since the shallow features are more scattered, as shown in \cref{sec:idea}.}

\eccvVersion{To effectively transfer discriminative cues among samples from shallow features to rectified features and avoid introducing noise, we propose two relation-based transfer losses at different levels.}
Firstly, we offer a \textbf{Instance-level Relation transfer} (IR) to mine the relation between feature instances.
Given a pair of final features $X_{N_L}^1$ and $X_{N_L}^2$, and another pair of intermediate features $X_{l}^1$ and $x_l^2$, the IR loss is defined as:
\begin{equation}
    \mathcal{L}_{IR} = \delta(\epsilon(X_{FR}^1, X_{FR}^2), \epsilon(X_{l}^1, x_l^2)),
\end{equation}
where $\epsilon(\cdot)$ is the Euclidean distance and the $\delta(\cdot)$ denotes the smooth L1 loss.
Besides the relation between feature instances, another level of the relation information is the relation between features and class centers.
\textbf{Class-center Relation Transfer} (CR) is proposed to mine this class-center-level information.
Assume at task $T_i$ with $|\mathcal{Y}_i|$ classes and the feature dimension is $d$.
Given a set of class prototypes $P \in \mathbb{R}^{d \times |\mathcal{Y}_i|}$  from the classifier $\Phi(\cdot)$  as well as a pair of final feature $X_{N_L} \in \mathbb{R}^{d \times 1}$ and the intermediate feature $X_l$ ($l\in[1,N_L)$), the corresponding CR loss is formulated as:
\begin{equation}
    \mathcal{L}_{CR} = \text{KL}((\sigma(P^\top\overline{X_{FR}}), \sigma(P^\top\overline{X_{l}})),
\end{equation}
where $\sigma(\cdot)$ is the softmax function, and KL$(\cdot)$ is the Kullback-Leibler divergence. $\overline{X}$ denotes the L2 normalization results of $X$.

In the above equations, the two losses $\mathcal{L}_{CR}$ and  $\mathcal{L}_{IR}$ work on different levels and serve different roles.
$\mathcal{L}_{CR}$ works on the relations between a sample and all class centers, focusing on the coarse-grained class-specific patterns that distinguish different classes.
Differently, $\mathcal{L}_{IR}$ provides knowledge about fine-grained discriminative cues that contribute to pairwise relations.
The proposed relation-based transfer losses are related to previous knowledge distillation works\cite{rkd, distill1, distill2}.
However, these works do not share the same objective as ours.
Our goal is to extract precious transferrable information hidden in various shallow layers hierarchically to rectify the final features while prior distillation methods focus on matching final outputs to teachers.
Furthermore, there exists a branch of works using intermediate features\cite{inter1,inter2} directly for dense prediction. Differently, we are the first to extract intermediate knowledge to rectify the final feature for a classification task.

Besides, we constrain the discriminative ability of rectified features on base classes to avoid negative rectification.
Since the final feature of layer $L_{N_L}$ is strong enough for the base classes, we apply a cosine constraint as:
\begin{equation}
    \mathcal{L}_{cos} = \text{cos}(X_{FR}, X_{N_L}),
\end{equation}
where cos$(\cdot, \cdot)$ is cosine similarity.

\noindent\textbf{Novel class training.}
To ensure the rectified features are centered around the class mean prototypes, we further introduce a classification loss for novel classes:
\begin{equation}
    \mathcal{L}_{NovCE} = - \mathbbm{1}(y \in \cup_{n>1} \mathcal{Y}_n) \log \sigma(P^\top\overline{X_{FR}})^{(y)},
\end{equation}
where $y$ is the corresponding ground true label of $X_{FR}$, $(P^\top\overline{X_{FR}})^{(y)}$ is the $y$-th element of $(P^\top\overline{X_{FR}})$.

\noindent\textbf{Anti-forgetting training.}
Another challenge of FSCIL is the forgetting problem.
Thanks to our feature rectification design, the only trainable module is the FR module.
As the FR module takes the final and the intermediate feature as the input, these kinds of features can be easily modeled by Gaussian distributions~\cite{PASS,apg}.
Hence, we construct a Gaussian distribution for each class on each layer.
The Gaussian distribution of class $i$ features at layer $L_l$ can be denoted as $\mathcal{N}_{i}^l(\mu_i^l, \Sigma_i^l)$, where $\Sigma_i^l$ is obtained by calculating the covariance of each feature dimension if $i \in \mathcal{Y}_1$ (base classes). 
For novel classes, the covariance matrix is computed by averaging the covariance matrices of the top-k most similar base classes.
During the incremental learning phases, we utilize these Gaussian distributions to sample features, which serve as inputs from old tasks to the FR module, assisting the FR module in retaining previously learned knowledge.

\noindent\textbf{Overall training objectives.}
The total loss function for our feature rectification framework is as follows:
\begin{equation}
    \mathcal{L} = \beta_{cos}(\mathcal{L}_{cos} + \mathcal{L}_{NovCe}) + \beta_{CR}\mathcal{L}_{CR} + \beta_{IR}\mathcal{L}_{IR},
\end{equation}
where $\beta_{cos}, \beta_{CR}$ and $\beta_{IR}$ are trade-off parameters.

\noindent\textbf{Multi-layer knowledge ensemble.}
Since there exists hierarchical knowledge in different layers, 
we mine such information by knowledge ensemble from different layers.
Specifically, we incorporate an FR module with each selected layer, resulting in a total of $n_{FR}$ FR modules. 
During the training phase, these FR modules are optimized simultaneously. 
When evaluation, FR modules at layer $l$ are employed to produce multiple rectified features denoted as $X^l_{FR}$. 
These rectified features are subsequently input into the shared
classifier $P$ for generating multiple predictions: Pred$_l = \sigma(P^\top\overline{X^l_{FR}})$.
The final prediction is computed as the average of each FR-branch: $\text{Pred} = \frac{1}{n_{FR}}\sum_{l=1}^{n_{FR}}\text{Pred}_l$.

\section{Experiments}
\vspace{-5pt}
\label{sec:exp}

\begin{table*}[pt] 
        \scriptsize
		\centering
  \captionsetup{font=scriptsize}
		\caption{Comparison with FSCIL methods on \emph{mini}ImageNet. 
            The best and second best results are marked \textbf{bolded} and \underline{underlined}.
            Methods with\dag are reported with the original results in their paper. Other results are reimplemented by us.\vspace{-8pt}}
        \resizebox{300pt}{!}{
			\begin{tabular}{lc|cccccccccccc}
				\toprule
				\multicolumn{1}{l}{\multirow{2}{*}{\bf Methods}} & \multicolumn{1}{c}{\multirow{2}{*}{\bf Num. of}} & \multicolumn{9}{c}{\bf $aAcc$ in each task (\%)} & \bf  &  \\ 
				\cmidrule{3-13}
				& \bf Params.(M) &\bf 1      & \bf 2      & \bf 3    & \bf 4     & \bf 5  & \bf 6     & \bf 7      &\bf 8  & \bf 9    & \bf $aAcc$ & $gAcc$ \\ 
				\midrule
				TOPIC~\cite{topic}\dag  & - &61.3	& 50.1	& 45.2	& 41.2	& 37.5	& 35.5	& 32.2	& 29.5	& 24.4	& 39.64 & -\\
				IDLVQ~\cite{chen2020incremental}\dag  & - &64.8 &	59.8 & 55.9	& 52.6	& 49.9	& 47.6	& 44.8	& 43.1	& 41.8	& 51.16 & - \\
                MetaFSCIL~\cite{metafscil}\dag & - &72.0 & 68.0 & 63.8 & 60.3 & 57.6 & 55.2 & 53.0 & 50.8 & 49.2 & 58.85 & -  \\
                DF-Replay~\cite{data_free}\dag  & - &71.8 & 67.1 & 63.2 & 59.8 & 57.0 & 54.0 & 51.6 & 49.5 & 48.2 & 58.02 & -  \\
                FCIL~\cite{FCIL}\dag  & - & 76.34 & 71.4 & 67.1 & 64.1 & 61.3 & 58.5 & 55.7 & 54.1 & 52.8 & 62.37 & -  \\
				Self-promoted~\cite{sppr}  & 12.34 & 63.6 & 64.1 & 59.5 & 55.5 & 52.1 & 49.2 & 45.6 & 43.1 & 40.9 & 52.62 & 41.53 \\
				CEC~\cite{cec}    & 12.12 &72.1	& 67.0 & 63.0 & 59.6 & 56.9 & 53.9 & 51.4 & 49.5 & 47.8 & 57.91 & 48.64  \\
				LIMIT~\cite{limit} & 12.29 &73.3 & 67.6 & 63.2 & 59.5 & 56.7 & 53.7 & 51.4 & 49.7 & 48.2 & 58.15 & 49.47  \\
                FACT~\cite{fact} & 11.26 &76.3 & 71.3 & 66.9 & 62.9 & 59.6 & 56.3 & 53.5 & 51.1 & 49.2 & 60.78 & 49.56  \\
                CLOM~\cite{clom} & 14.22 &72.1 & 66.7 & 62.7 & 59.2 & 56.1 & 53.0 & 50.2 & 48.3 & 46.7 & 57.23 & 46.11  \\
				C-FSCIL~\cite{cfscil}  & 12.79 &70.6 & 63.8 & 59.3 & 56.4 & 54.3 & 51.8 & 48.9 & 47.3 & 46.1 & 55.40 & 50.46  \\
                BiDistFSCIL~\cite{BiDistFSCIL} & 11.50 &74.6 & 70.4 & 66.6 & 62.8 & 60.5 & 57.2 & 54.7 & 53.0 & 51.9 & 61.32 & 53.49  \\
                S3C~\cite{s3c} & 0.34 &77.0 & 72.1 & 68.2 & 65.1 & 62.0 & 58.8 & 56.0 & 53.6 & 51.9 & 62.74 & 53.70  \\
                Regularizer~\cite{regularizer}  & 26.30 &81.9 & 72.7 & 68.7 & 64.1 & 57.4 & 53.5 & 53.8 & 52.3 & 43.1 & 62.82 & 55.62  \\
                SubNetwork~\cite{subnetwork} & 22.38 &78.1 & 73.5 & 69.8 & 66.2 & 63.2 & 60.3 & 57.6 & 55.7 & 54.5 & 64.31 & 55.88  \\
                SAVC~\cite{savc} & 12.12 &81.0 & 76.2 & 71.9 & 71.4 & 65.6 & 62.1 & 59.2& 57.0 & 55.5 & 66.66 & 56.35  \\
                ALICE\cite{alice} & 11.20 &81.1	& 71.1	& 67.8	& 64.1	& 62.4	& 60.1	& 57.9	& 56.8	& 56.0	& 64.14 &  59.72  \\
				NC-FSCIL\cite{neural_collapse} & 15.87 &84.2 &	76.1 &	72.1 &	68.1 & 67.3 & 63.9&	61.9& 59.6 & \underline{57.9} & \underline{67.90} & \underline{60.93}  \\
			\midrule
                YourSelf (Ours) & 11.57 &84.0 & 77.6 & 73.7 & 70.0 & 68.0 & 64.9 & 62.1 & 59.8 & \textbf{59.0} & \textbf{68.80} & \textbf{62.20} \\
				\bottomrule
			\end{tabular}
   }
		\label{table:imgnet}
\vspace{-20pt}
\end{table*}

\subsection{Experiment Setup}
\noindent\textbf{Datasets.} Following previous works, we conduct experiments on three popular datasets of FSCIL: CIFAR-100~\cite{cifar100}, \emph{mini}ImageNet~\cite{mini} and CUB-200~\cite{cub200}. For statistics of the three datasets, please refer to the supplementary material.

\noindent\textbf{Training and testing protocols.} For training protocols, we follow prior works~\cite{cfscil,BiDistFSCIL,s3c,regularizer,subnetwork,savc,alice,neural_collapse} to split a large proportion of the training set to serve as the base task, and split the rest into small incremental tasks. 
Specifically, for CIFAR-100 and \emph{mini}ImageNet, the first task contains 60 classes out of the total 100 classes, and the rest 40 classes are split into 8 tasks, following a 5-way 5-shot scheme.
For CUB-200, there are 100 classes in the first task and the number of novel tasks is 10 (in a 10-way 5-shot setting).
As for the testing, we follow prior works to report the \textit{average accuracy (aAcc)}. In addition, we also provide detailed discussions on our metric \textit{generalized accuracy (gAcc)}.
\eccvVersion{We further provide detailed results of novel classes only in our supplementary material.}

\noindent\textbf{Implementation details.} 
For all experiments below, we use the ViT suggested by Touvron et, al.~\cite{deit}. 
\eccvVersion{We adjust stuff like the dimensions of the backbone to match the number of parameters (around 10M) as the ResNet18~\cite{resnet} which is widely used in this topic. 
The Feature Rectification (FR) is a lightweight module and uses popular MLPs implementation~\cite{deit} with only 0.6M parameters.
To ensure a fair comparison, we also re-implement other methods with the same backbone as ours, but found the performance is rarely improved. 
Please refer to Supplementary material for more details.}
We follow previous works to \textbf{train from scratch} on CIFAR-100 and \emph{mini}ImageNet dataset.
Prior works~\cite{topic,chen2020incremental,metafscil,data_free,sppr,cec} adopt an ImageNet-pretrained ResNet for the CUB-200 dataset.
To make a fair comparison, we also pretrained our backbone on the ImageNet~\cite{imagenet} to evaluate on CUB-200.
For evaluation, we reproduced 14 methods based on the public codes and conducted in-depth comparisons in the following sections.

\begin{figure*}[t]
\begin{minipage}{0.49\textwidth}
    \centering
    \captionsetup{font=scriptsize}
      \includegraphics[width=1.02\linewidth]{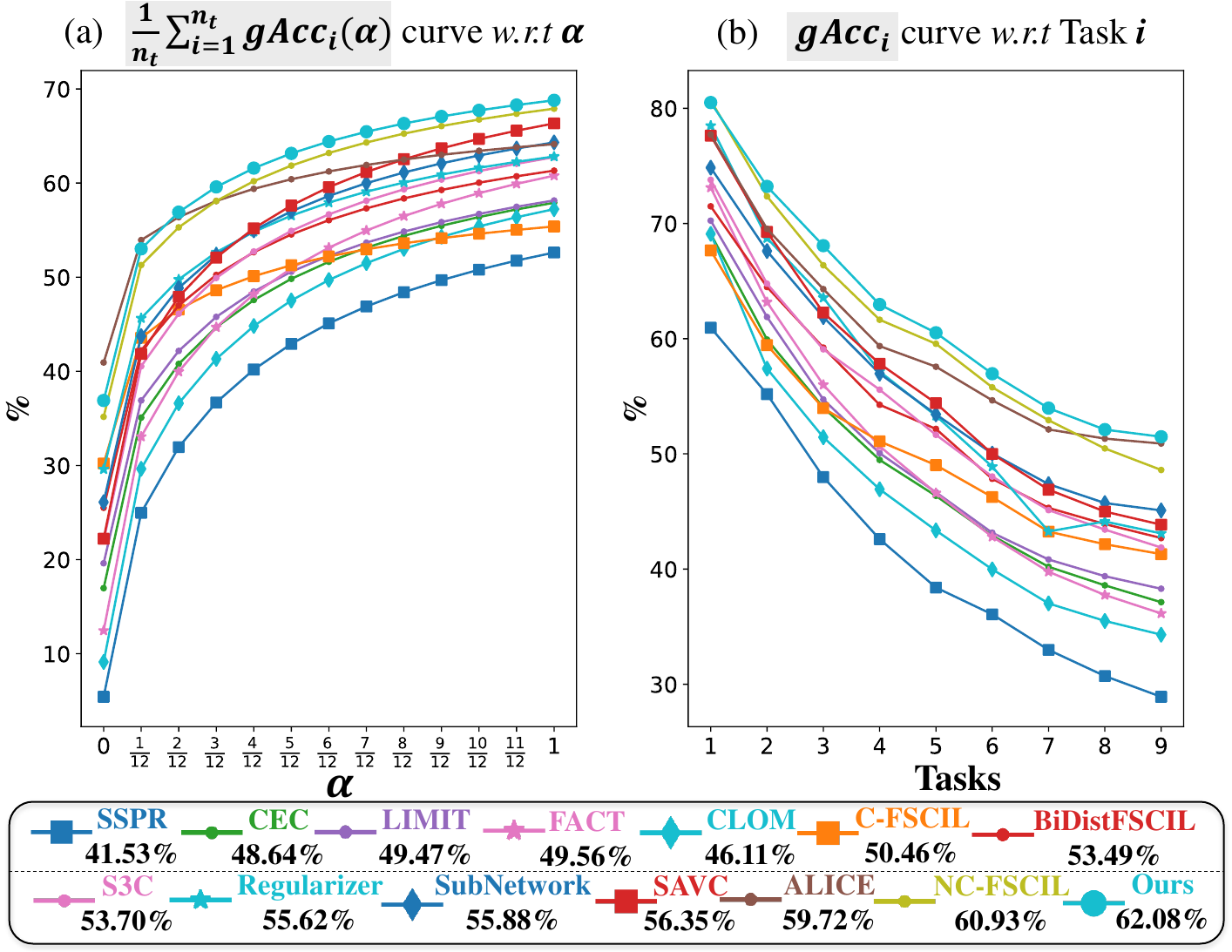}
      \caption{
      FSCIL results on \emph{mini}ImageNet.
      (a): The $gAcc$ curve (averaged across all $n_t$ tasks) \vs $\alpha$.
      (b): The $gAcc$ AUC of each task $\mathcal{T}_i$ (\cref{eq:gacci})\vs $i$.
      In the legend, we show the AUC value (\cref{eq:gacc}) of each method.
      Best viewed in color.
      \vspace{-18pt}
      }
    \label{fig:exp_mini_all}
  \end{minipage}
  \hfill
\begin{minipage}{0.49\textwidth}
    \centering
    \captionsetup{font=scriptsize}
      \includegraphics[width=\linewidth]{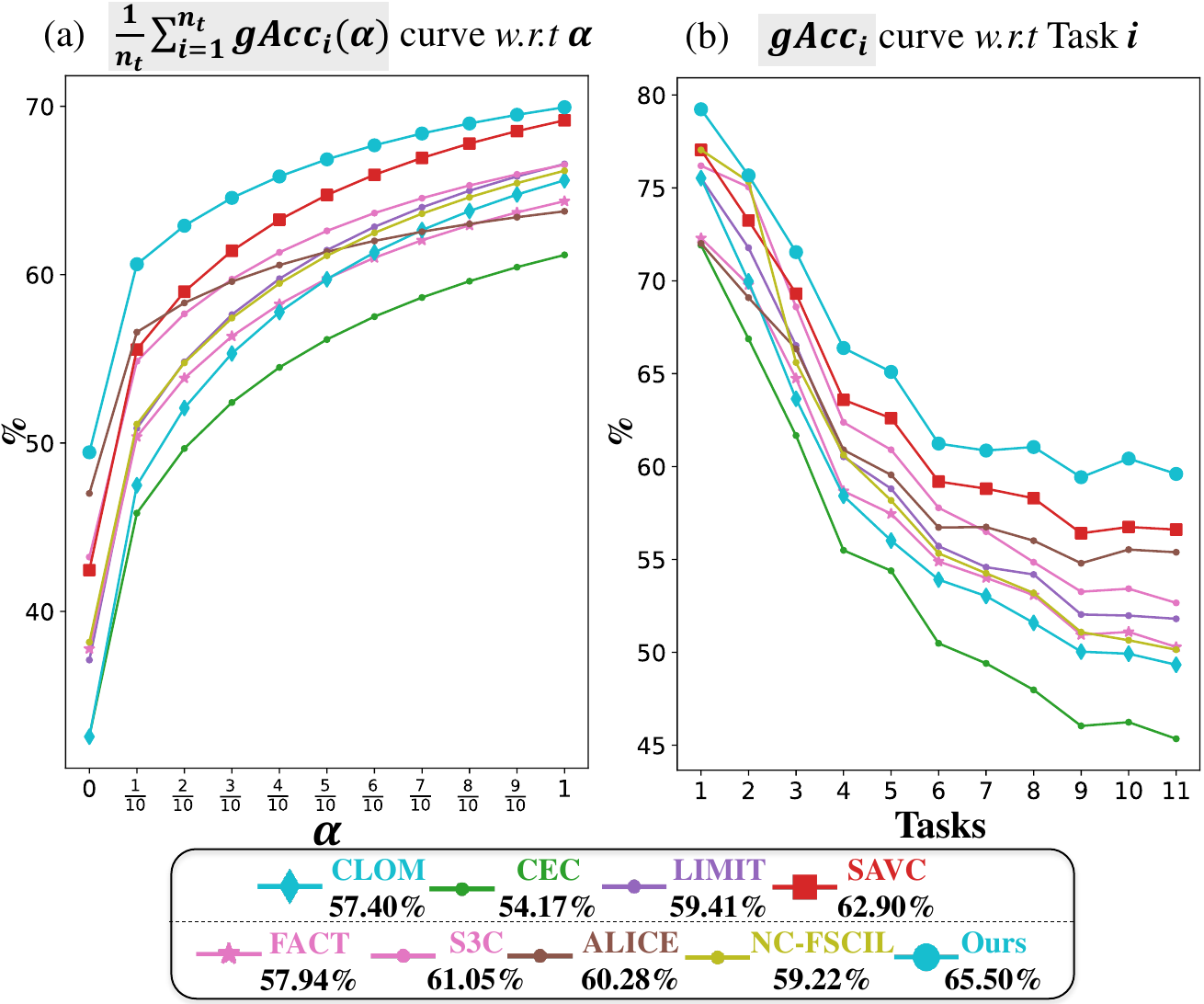}
      \caption{
      FSCIL results on CUB-200.
      (a): The $gAcc$ curve (averaged across all $n_t$ tasks) \vs $\alpha$.
      (b): The $gAcc$ AUC of each task $\mathcal{T}_i$ (\cref{eq:gacci})\vs $i$.
      In the legend, we show the AUC value (\cref{eq:gacc}) of each method.
      Best viewed in color.
      \vspace{-18pt}
      }
    \label{fig:exp_cub_all}
  \end{minipage}
\end{figure*}

\subsection{Comparison with State-of-the-art methods}
Experimental results on \emph{mini}ImageNet are shown in \Cref{table:imgnet}.
We report the classical accuracy ($aAcc$) in each section as well as its averaged results and our proposed $gAcc$.
From the table, we can conclude that our proposed method outperforms other methods by a large margin.
Specifically, under the classical evaluation metric $aAcc$, the proposed method achieves 68.80\% accuracy, which is 0.90\%, 4.66\%, 2.14\%  higher than SOTA methods NC-FSCIL~\cite{neural_collapse}, ALICE~\cite{alice} and SAVC~\cite{savc}.
The improvement is non-trivial considering that \emph{mini}ImageNet is a complex dataset.

From a more balanced perspective, we also report $gAcc$ on the \Cref{table:imgnet} and \cref{fig:exp_mini_all}.
In the left figure of \cref{fig:exp_mini_all}, except for the method ALICE when $\alpha = 0$, our method outperforms most other methods by a clear margin\eccvVersion{(at leat 1.27\%)} and achieves the highest AUC of all methods.
This indicates that our method achieves a more balanced base-novel performance.
To see raw numbers of the accuracy of novel classes, please refer to the supplementary material.

Besides, \cref{fig:exp_mini_all} shows that methods that perform well at $\alpha = 1$ do not necessarily perform well when $\alpha$ is smaller.
For example, although CLOM achieves 57.23\% average accuracy ($\alpha = 1$) and is 1.83\% higher than method C-FSCIL (55.4\%), we can see the performance drops significantly when $\alpha$ is small.
It is mainly because CLOM focuses on the base-classes accuracy and performs poorly on novel classes.
We can observe that the AUC of the CLOM is much smaller (4.35\%) than that of C-FSCIL.
The results confirm the necessity of our proposed $gAcc$.
Similar phenomena can be found when comparing other methods like SAVC and ALICE, and we can obtain similar conclusions on the other two datasets including CUB-200 and CIFAR-100.

Results on CUB-200 dataset are shown in \Cref{table:cub200} and \cref{fig:exp_cub_all}.
We can see from \Cref{table:cub200} that our method is better than any other existing methods in terms of $aAcc$ and $gAcc$ (1.52\%@$aAcc$ and 2.6\%@$gAcc$ improvement over the strong method SAVC).
Results on CIFAR-100 dataset are shown in \Cref{table:cifar}.
For the traditional metric $aAcc$, our method achieves comparable performance with SOTA methods, while our method outperforms all other methods by a clear margin when considering the newly introduced metric $gAcc$.
Please refer to Sec \textcolor{red}{A5} of the supplementary material for more experiment results.

\begin{table*}[t] 
    \scriptsize
	\centering
        \captionsetup{font=scriptsize}
            \caption{Comparison with FSCIL methods on CUB-200. 
            The best and second best results are marked \textbf{bolded} and \underline{underlined}.
            Methods with~\dag~are reported with the original results in their paper. Other results are reimplemented by us.
            \vspace{-10pt}
            }
        \resizebox{300pt}{!}{
			\begin{tabular}{lc|ccccccccccccc}
				\toprule
				\multicolumn{1}{l}{\multirow{2}{*}{\bf Methods}}& \multicolumn{1}{c}{\multirow{2}{*}{\bf Num. of}} & \multicolumn{11}{c}{\bf $aAcc$ in each task (\%)} & \bf  &  \\ 
				\cmidrule{3-12}
				& \bf Params.(M) & \bf 1      & \bf 2      & \bf 3    & \bf 4     & \bf 5  & \bf 6     & \bf 7      &\bf 8   & \bf 9  &\bf 10 & \bf 11 &\bf  $aAcc$ & $gAcc$ \\ 
				\midrule
				TOPIC~\cite{topic}\dag  & - & 68.7	& 42.5 & 54.8 & 50.0 & 45.3 & 41.4 & 38.4 & 35.4 & 32.2 & 28.3 &  26.3 & 43.92 & - \\
                DF-Replay~\cite{data_free}\dag & - & 75.9 & 72.1  & 68.6 & 63.8 & 62.6 & 59.1 & 57.8 & 55.9 & 55.0 & 53.6 & 52.4 & 61.52 & - \\
                BiDistFSCIL~\cite{BiDistFSCIL}\dag & - & 79.1 & 75.4 & 72.8 & 69.1 & 67.5 & 65.1 & 64.0 & 63.5 & 61.9 & 61.5 & 60.9 & 67.34 & - \\
                MetaFSCIL~\cite{metafscil}\dag & - & 75.9 & 72.4 & 68.8 & 64.8 & 63.0 & 60.0 & 58.3 & 56.8 & 54.8 & 53.8 & 52.6 & 61.93 & - \\
                SubNetwork~\cite{subnetwork}\dag & - & 78.7 & 74.6 & 71.4 & 67.5 & 65.4 & 62.6 & 61.1 & 59.4 & 57.5 & 57.2 & 56.7 & 64.68 & - \\
                DSN~\cite{dsn}\dag & - & 76.1 & 72.2 & 69.6 & 66.7 & 64.4 & 62.1 & 60.2 & 58.9 & 57.0 & 55.1 & 54.2 & 63.31 &- \\
                Self-promoted~\cite{sppr} & 12.36 & 62.6 & 54.7 & 51.6 & 47.3 & 44.3 & 40.9 & 38.8 & 39.0 & 35.9 & 35.1 & 33.5 &  44.07  &  40.88\\
				CEC~\cite{cec}  & 12.34 & 75.7 & 71.4 & 68.1 & 64.0 & 62.0 & 58.8 & 57.3 & 55.5 & 53.6 & 53.2 & 52.1 & 61.17 & 54.16 \\
				LIMIT~\cite{limit} & 12.33 & 79.5 & 76.4 & 73.2 & 69.2 & 67.1 & 64.3 & 62.8 & 61.9 & 59.8 & 59.1 & 57.6 & 66.58 & 59.40 \\
                CLOM~\cite{clom} & 18.89 & 79.5 & 75.9 & 72.0 & 68.2 & 65.7 & 63.4 & 62.0 & 60.3 & 58.5 & 57.9 & 56.9 & 65.61 & 57.40 \\
                FACT~\cite{fact} & 11.34 & 76.1 & 73.4 & 70.4 & 66.5 & 64.8 & 62.5 & 61.1 & 59.9 & 57.9 & 57.5 & 56.5 & 50.29 & 57.93 \\ 
                NC-FSCIL~\cite{neural_collapse} & 15.91 & 81.1 & 76.4 & 72.8 & 69.3 & 66.1 & 63.7 & 62.2 & 61.0 & 59.0 & 58.0 & 57.3 & 66.17 & 59.22 \\
                ALICE~\cite{alice} & 41.71 & 75.8 & 70.1 & 68.4 & 65.2 & 63.7 & 61.2 & 60.7 & 59.7 & 58.5 & 58.7 & 58.4 & 63.77 & 60.28 \\
                S3C~\cite{s3c} & 13.06 & 80.2 & 76.5 & 72.9 & 69.1 & 66.9 & 64.3 & 62.7 & 61.0 & 59.5 & 59.2 & 58.3 & 66.53 & 61.05 \\
                SAVC~\cite{savc} & 24.29 & 81.1 & 77.6 & 75.0 & 71.2 & 69.6 & 66.7 & 65.8 & 64.8 & 62.9 & 62.7 & \underline{62.2} & \underline{68.33} & \underline{62.90} \\
				\midrule
                YourSelf (Ours) & 11.90 & 83.4 & 77.0 & 75.3 & 72.2 & 69.0 & 66.8 & 66.0 & 65.6 & 64.1 & 64.5 & \textbf{63.6} & \textbf{69.85} & \textbf{65.50} \\
				\bottomrule
			\end{tabular}
   }
   \vspace{-10pt}
		\label{table:cub200}
\end{table*}

 \begin{table}[t]
    \scriptsize
	\centering
  \captionsetup{font=scriptsize}
  		\caption{Comparison results on CIFAR-100. 
            The best and second best results are marked \textbf{bolded} and \underline{underlined}. 
            All results of other methods are reimplemented by us.
            See more results in the supplementary material.
            \vspace{-10pt}
            } 
  \resizebox{260pt}{!}{
            \renewcommand\arraystretch{1}
			\begin{tabular}{lc|ccccccccccc}
				\toprule
				\multicolumn{1}{l}{\multirow{2}{*}{\bf Methods}} & \multicolumn{1}{c}{\multirow{2}{*}{\bf Num. of}}& \multicolumn{9}{c}{\bf $aAcc$ in each task (\%)} & \bf  &  \\ 
				\cmidrule{3-11}
				&\bf Params.(M) & \bf 1      & \bf 2      & \bf 3    & \bf 4     & \bf 5  & \bf 6     & \bf 7      &\bf 8 & \bf 9    & \bf $aAcc$ & $gAcc$ \\ 
				\midrule
								CEC~\cite{cec}  & 0.296 &  73.2 & 69.1 & 65.4 & 61.2 & 58.0 & 55.5 & 53.2 & 51.3 & 49.2 & 59.56 & 50.85  \\
				LIMIT~\cite{limit}  & 0.296 &  76.1 & 71.7 & 67.3 & 63.2 & 60.1 & 57.3 & 55.2 & 53.0 & 50.8 & 61.64 & 52.16  \\
                CLOM~\cite{clom} & 0.355 & 74.2 & 69.8 & 66.1 & 62.2 & 59.0 & 56.1 & 54.0 & 51.9 & 49.8 & 60.35 & 50.36  \\
				C-FSCIL~\cite{cfscil} & 12.79 & 76.8 & 72.0 & 67.2 & 63.1 & 59.7 & 56.7 & 54.4 & 51.9 & 49.7 & 61.28 & 50.62  \\
                SAVC~\cite{savc} & 0.842 &  78.4 & 72.7 & 68.4 & 64.0 & 61.2 & 58.3 & 56.0 & 54.1 & 51.7 & 62.76 & 53.29  \\
                FACT~\cite{fact} & 0.282 &  78.5 & 72.7 & 68.8 & 64.5 & 61.3 & 58.6 & 56.8 & 54.4 & 52.3 & 63.12 & 54.56  \\ 
                SubNet~\cite{subnetwork}  & 22.39 & 80.0 & 75.7 & 71.6 & 67.8 & 64.7 & 61.4 & 59.3 & 57.1 & 55.1 & 64.85 & 56.63 \\
                S3C~\cite{s3c} & 0.341 &  78.0 & 73.9 & 70.2 & 66.1 & 63.3 & 60.1 & 58.3 & 56.7 & 54.0 & 64.51 & 58.04  \\
                ALICE~\cite{alice} & 45.28 &  80.3	& 71.9	& 67.0	& 63.2	& 60.7	& 58.3	& 57.3	& 55.5	& 53.7	& 63.10 & 59.72 \\
                NC-FSCIL~\cite{neural_collapse} & 15.93 & 82.9	& 77.5	& 73.7	& 69.0	& 65.6 & 62.0 & 59.7 & 57.8 & \underline{55.4}	& \textbf{67.06} & \underline{59.73} \\
				\midrule
				Ours & 11.68  &  82.9	& 76.3	& 72.9	& 67.8	& 65.2	& 62.0	& 60.7	& 58.8	& \textbf{56.6}	& \underline{67.02} & \textbf{60.32}\\
				\bottomrule
			\end{tabular}
   }
      \vspace{-20pt}
		\label{table:cifar}
\end{table}

\subsection{Ablation studies}

\begin{figure*}[t]
\hspace{-17pt}
  \begin{minipage}{0.59\textwidth}
    \begin{minipage}{\textwidth}
    \scriptsize
    \captionsetup{font=scriptsize}
      \captionof{table}{Ablation studies on CIFAR-100.}
    \resizebox{218pt}{!}{
      \renewcommand\arraystretch{1.0}
        \begin{tabular}{ccccccccccccccc}
        \toprule
        \multirow{2}*{{FR}} & \multirow{2}*{$\mathcal{L}_{Cos}$\ + $\mathcal{L}_{NovCE}$} & \multirow{2}*{{$\mathcal{L}_{CR}$}} & \multirow{2}*{\textbf{$\mathcal{L}_{IR}$}} & \multicolumn{9}{c}{aAcc. in each task (\%)} & \multirow{2}{*}{$aAcc$} & \multirow{2}{*}{$gAcc$}  \\
    \cline{5-13}                &       &       &      & 1     & 2     & 3     & 4     & 5     & 6     & 7     & 8  &  9 \\
        \hline
              &       &             & & 82.9 & 77.8 & 72.9 & 68.4 & 64.6 & 61.4 & 58.7 & 56.4 & 54.1 & 66.35 & 54.20  \\
        \checkmark     & \checkmark  &  & & 82.9 & 76.7 & 73.1 & 67.9 & 64.8 & 60.8 & 59.2 & 57.2 & 53.9 & 66.26 & 57.48  \\
        \checkmark     & \checkmark        & \checkmark     &   & 82.9 & 76.1  & 72.4  & 67.8 & 64.5 & 61.3 & 59.6 & 57.3 & 54.9 & 66.33 & 58.55\\
        \checkmark     & \checkmark         & \checkmark     &\checkmark    &  82.9	& 75.7	& 72.5	& 67.4	& 64.8	& 61.9	& 60.4	& 58.7	& 56.3	& 67.02 & 60.32  \\
        \bottomrule
        \end{tabular}%
        }
        \vspace{-10pt}
        \label{tab:ablate_cifar}
       \end{minipage}
  
  \begin{minipage}{\textwidth}
    \scriptsize
    \captionsetup{font=scriptsize}
      \captionof{table}{Ablation studies on \emph{mini}ImageNet.}
    \resizebox{218pt}{!}{
      \renewcommand\arraystretch{1}
        \begin{tabular}{ccccccccccccccc}
        \toprule
        \multirow{2}*{{FR}} & \multirow{2}*{$\mathcal{L}_{Cos}$\ + $\mathcal{L}_{NovCE}$} & \multirow{2}*{{$\mathcal{L}_{CR}$}} & \multirow{2}*{\textbf{$\mathcal{L}_{IR}$}} & \multicolumn{9}{c}{aAcc. in each task (\%)} & \multirow{2}{*}{$aAcc$} & \multirow{2}{*}{$gAcc$}  \\
    \cline{5-13}                &     &       &     & 1     & 2     & 3     & 4     & 5     & 6     & 7     & 8  & 9  \\
        \hline
              &       &             & & 84.0 & 79.4 & 75.0 & 71.1 & 67.8 & 64.4 & 61.3 & 58.8 & 56.9 & 68.74 & 57.87  \\
        \checkmark     & \checkmark    &  & & 84.0 & 74.0 & 69.7 & 66.8 & 66.1 & 63.5 & 60.8 & 58.6 & 58.1 & 66.84 & 60.89  \\
        \checkmark     & \checkmark        & \checkmark     &   & 84.0 & 77.7   & 73.6   & 69.9  & 67.4  & 64.6 & 61.6  & 59.8  & 58.7 & 68.60 & 61.38 \\
        \checkmark     & \checkmark         & \checkmark     &\checkmark    &84.0 & 77.6 & 73.7 & 70.0 & 68.0 & 64.9 & 62.1 & 59.8 & 59.0 & 68.80 & 62.20 \\
        \bottomrule
        \end{tabular}%
        }
        \label{tab:ablate_mini}
       \end{minipage}
    \end{minipage}
\hfill
\hspace{13pt}
  \begin{minipage}{0.45\textwidth}
  \begin{minipage}{\textwidth}
  \vspace{15pt}
    \includegraphics[width=\linewidth]{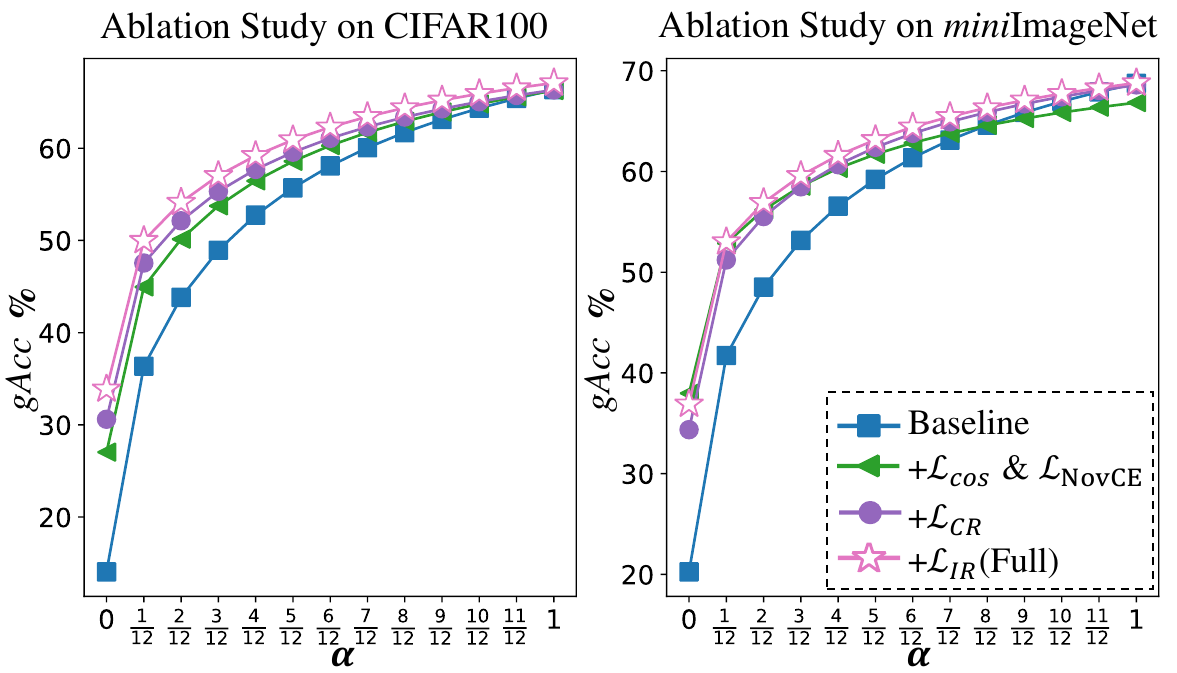}
     \vspace{-20pt}
     \captionsetup{font=scriptsize}
    \caption{Ablation studies on two datasets, evaluating using $gAcc$.}
    \label{fig:exp_ablation_all}
  \end{minipage}
  \end{minipage}
    \vspace{-20pt}
\end{figure*}

\noindent\textbf{Effectiveness of different losses.}
Ablation experiments of the effectiveness of $\mathcal{L}_{IR}$ and $\mathcal{L}_{CR}$ are conducted among CIFAR-100, \emph{mini}ImageNet, and CUB-200 datasets.
Experimental results are shown in \Cref{tab:ablate_cifar} and \Cref{tab:ablate_mini}.
The detailed $gAcc$ \vs $\alpha$ as shown in \cref{fig:exp_ablation_all}.
Results on CIFAR-100 show that introducing the FR module can effectively enhance the $gAcc$ by 3.28\%.
After further introducing the class-relation (CR) transfer loss, the performance at $gAcc$ is further improved by 1.07\%.
Adding the instance-relation (IR) transfer loss will further boost the $gAcc$ by 1.91\%.
The effectiveness of the proposed losses is also verified on the \emph{mini}ImageNet dataset.
For \emph{mini}ImageNet dataset, the introduction of FR modules brings 3.02\% improvement at $gAcc$.
Applying loss $\mathcal{L}_{CR}$ and $\mathcal{L}_{IR}$ can improve $gAcc$ by 0.49\% and 0.82\%, respectively.
It is worth mentioning that $\mathcal{L}_{CR}$ and $\mathcal{L}_{IR}$ can also boost the $aAcc$.
Specifically, on the \emph{mini}ImageNet, 
adding these two losses results in an increase at $aAcc$ performance of 1.96\%.
On CIFAR-100, the $aAcc$ is improved by about 0.76\% with the help of these two losses.
See more experiments in the supplementary material.

\noindent\textbf{Effectiveness of multi-layer knowledge ensemble.}
In \cref{sec:method}, we propose to exploit hierarchical knowledge from multiple layers for effective feature rectification.
In practice, we select layers 8, 9 and 10 for the ensemble. 
Here, we take a more detailed discussion on each layer before the ensemble of multiple layers.
We conducted experiments on the \emph{mini}ImageNet, and the results are shown in \Cref{tab:ablate_mini_layer}.
It is clear that for each layer, the corresponding FR module effectively calibrates the features and achieves improvement from 2.23\% to 3.54\% at $gAcc$.
Furthermore, the introduction of the FR module does not harm the $aAcc$.
Considering the fact that different layers contain hierarchical knowledge, the multi-layer ensemble is proposed.
The results in the table show that the ensemble is effective in improving the $gAcc$ while maintaining $aAcc$.

\begin{table}[htbp]
  \centering
  \scriptsize
  \captionsetup{font=scriptsize}
  \caption{
    Detailed performance of each layer.
    Results are from \emph{mini}Imagenet dataset.
    `$l$-Raw': vanilla ViT features from layer $L_l$.
    \colorbox{myYellow}{`$l$-Recti'}: Rectified features using the proposed FR module. 
    \colorbox{myBlue}{`Ensemble'}: Ensembled results from `$l$-Recti', $l\in \{8,9,10\}$.
  }
  \renewcommand\arraystretch{1.2}
      \setlength{\tabcolsep}{1.1mm}
\resizebox{240pt}{!}{
    \begin{tabular}{cccccccccccccc}
    \toprule
    \multirow{2}*{\textbf{Layers}} & \multicolumn{9}{c}{$aAcc$ in each session (\%) } & \multirow{2}{*}{$aAcc$} & \multirow{2}{*}{$gAcc$} \\
\cline{2-10}          & 0     & 1     & 2     & 3     & 4     & 5     & 6     & 7     & 8 \\
    \hline
    8-Raw & 78.9 & 73.2 & 69.1 & 65.4 & 62.9 & 60.0 & 57.3 & 55.2 & 54.0 & 64.01 & 56.61 \\
    \rowcolor[HTML]{FFEFD5}8-Recti & 80.7 & 73.4  & 70.1  & 66.6  & 64.4  & 61.5  & 59.3  & 57.0  & 56.5  & 65.52  & 60.15  \\
    \hdashline
    9-raw & 82.5 & 77.3 & 73.2 & 69.7 & 65.3 & 63.7 & 60.9 & 58.9 & 57.5 & 67.66 & 59.33 \\
    \rowcolor[HTML]{FFEFD5}9-Recti & 83.0 & 76.6 & 73.2 & 69.7 & 67.5 & 64.6 & 61.9  & 59.9 & 59.0 & 68.39 & 61.90    \\
    \hdashline
    10-raw & 83.8 & 78.6 & 74.4 & 70.7 & 66.4 & 64.3 & 61.5 & 59.3 & 57.7 & 68.55 & 58.90 \\
    \rowcolor[HTML]{FFEFD5}10-Recti & 84.3 & 78.2 & 74.1 & 70.4 & 68.2 & 64.8 & 61.9  & 59.5 & 58.4 & 68.84 & 61.13  \\
    \hdashline
    \rowcolor[HTML]{E7EFFC}Ensemble &84.0 & 77.6 & 73.7 & 70.0 & 68.0 & 64.9 & 62.1 & 59.8 & 59.0 & 68.80 & 62.20\\
    \bottomrule
    \end{tabular}%
    }
  \label{tab:ablate_mini_layer}%
\end{table}%

\section{Conclusion}
\label{sec:conclusion}
In this work, we have revisited the challenges of the FSCIL, showing the limitation of the existing metric and introducing a novel metric denoted as \textit{generalized accuracy} ($gAcc$) for the comprehensive evaluation of FSCIL methods from a more balanced perspective. 
Furthermore, we incorporate the ViT into the FSCIL framework and unlock the potential of the ViT intermediate layer features. 
To achieve this, we have designed a feature rectification (FR) module aimed at integrating the valuable information from the intermediate layer features into the rectified feature space.
Using the new metric $gAcc$, we analyze the existing open-sourced FSCIL methods from a more balanced perspective and show the superiority of our method on three popular datasets.
\section*{Acknowledgements}
This work was supported partially by the National Key Research and Development Program of China (2023YFA1008503), NSFC(U21A20471, 62206315), Guangdong NSF Project (No. 2023B1515040025, 2020B1515120085, 2024A1515010101), Guangzhou Basic and Applied Basic Research Scheme(2024A04J4067).

\bibliographystyle{splncs04}
\bibliography{main}

\clearpage
\newpage
\section*{Appendices}
\renewcommand\thetable{A\arabic{table}}  
\renewcommand\thefigure{A\arabic{figure}} 
\renewcommand\thesection{A\arabic{section}} 
\renewcommand\theequation{A\arabic{equation}}
\setcounter{equation}{0}
\setcounter{section}{0}
\setcounter{table}{0}
\setcounter{figure}{0}
\section{Detailed Information of The Corner Cases}
\label{sec:corner_cases}
\begin{figure*}[h]
    \centering
      \includegraphics[width=\linewidth]{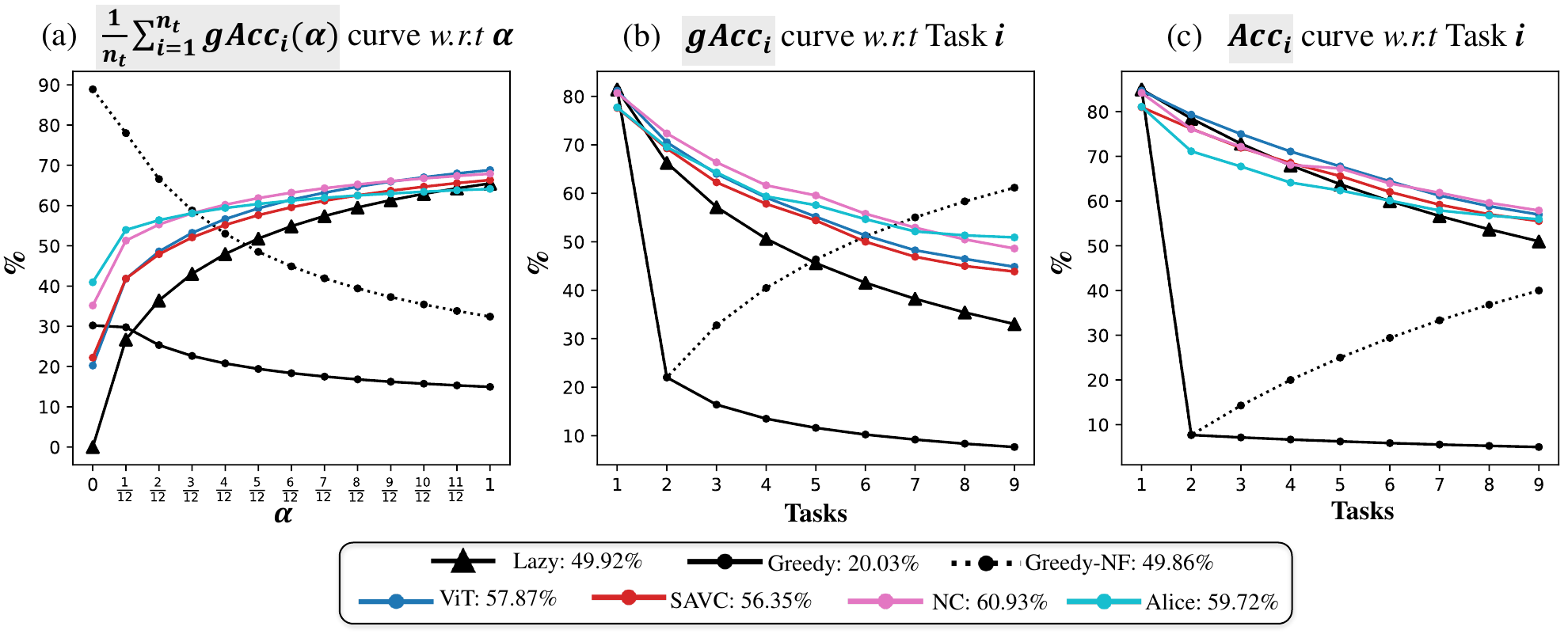}
      \caption{
      Detailed illustration of our proposed $gAcc$.
      This figure is an extension of the Figure \textcolor{red}{2} in the main paper.
      Performance of methods is obtained on the \emph{mini}ImageNet.
      In the legend, we show the average AUC across tasks (Eq. (\textcolor{red}{6)} in the main paper) of each method.
      \textbf{(a)}: The $gAcc$ curve (averaged across all $n_t$ tasks) v.s. param $\alpha$.
      \textbf{(b)}: The $gAcc$ AUC of each task $\mathcal{T}_i$ (see Eq. (\textcolor{red}{5)}) in the main paper) at each task.
      \textbf{(c)}: The $aAcc_i$ (see Eq. (\textcolor{red}{1}) in the main paper) at each task.
      Best viewed in color.
      }
    \label{fig:gacc_supp}
  \end{figure*}

In Sec 3. of the main paper, we analyze the necessity of the proposed \textit{generalized accuracy} ($gAcc$) metric.
In Fig. \textcolor{red}{2} of the main paper, we analyze some possible corner cases of the FSCIL setting on the proposed $gAcc$ metric.
Detailed information on these cases is shown below.
An extension of Fig. \textcolor{red}{2} of the main paper is shown in \cref{fig:gacc_supp}.

\noindent\textbf{Lazy}. To analyze the situation of an algorithm that totally ignores the novel tasks and only retrains the first-task (base) performance, we design this corner case called `Lazy' to show its performance under existing metrics and our proposed $gAcc$.
The detailed accuracies of each task are shown in \Cref{table:lazy}.
We set the base class performance as 85\% after training on the first task, \ie$A_1^1 = 85\%$.
We set the model completely \textbf{ignoring} the novel task and base task performance remains \textbf{unchanged} during incremental tasks, \ie $\forall j \geq i>1, A^j_i=0, A_i^1 = A_1^1$.
In \cref{fig:gacc_supp} (c), the accuracy ($aAcc$) of `Lazy' is very close to those of recent SOTAs.
This is not satisfying since the primary goal of continual learning is to learn new knowledge.
In comparison, the proposed $gAcc$ correctly distinguishes this situation and presents a more balanced evaluation, which gives a low AUC to `Lazy' and shows the margin of this case and other methods (see \cref{fig:gacc_supp} (b)).
In conclusion, the $aAcc$ alone fails to reflect the performance of FSCIL methods properly and our $gAcc$ can provide a more balanced perspective.

\noindent\textbf{Greedy and Greedy-NF}. We also show two other extreme examples. The `Greedy' situation is that an algorithm only focuses on the novel classes and does not care about the previous forgetting problem, \ie for all novel tasks, $A_i^j = 100\%$, when $i = j$ else $0$.
The $A^1_1$ is set to $ 85\%$.
See \Cref{table:greedy} for the detailed performance.
It is clear that, under the metric of $gAcc$ and $aAcc$, the 'greedy' performs far worse than other methods/cases.
Furthermore, we also show the non-forget version of the `Greedy' as `Greedy-NF'.
For 'Greedy-NF', we set $A_i^j$ to $100\%$ for $\forall j > 1,  j<i $.
See \Cref{table:greedy_nf} for the detailed performance at each task.
From \cref{fig:gacc_supp}, we can see that, after removing the forgetting problem of the `Greedy', the performance under $gAcc$ is boosted significantly.
It should be clarified that even though the 'Greedy-NF' performs well on the novel classes, their lame performance on the base classes affects the final $gAcc$.
These results show that the $gAcc$ will not over-emphasize the novel class performance and provide more balanced perspectives of base-novel performance.

\noindent\textbf{Vit baseline.}
We further show our vit baseline in the \cref{fig:gacc_supp}.
This baseline is conducted by employing a pure vit trained on the first task to extract features for novel classes.
This experiment is done on \emph{mini}ImageNet dataset.
Under the $gAcc$ metric, it is shown that even though the vit baseline gets a high performance on $gAcc(1)$, it does not perform well on the novel classes thus the AUC of the baseline is worse than SOTA methods. 

\begin{table*}[t]
\setlength{\tabcolsep}{1mm}
\centering
\scriptsize
\caption{Detailed accuracies setting of the corner case \textbf{Lazy}.
In this case, the model only focuses on the base task and learns nothing about the novel tasks.
$f_i$: model after trained on task $\mathcal{T}_i$.
$A^i$: accuracy on task $\mathcal{T}_i$ of the corresponding model $f$.}
\label{table:lazy}
\begin{tabular}{c|cccccccccccc}
\midrule
(\%)& $A^1$ & $A^2$ & $A^3$ & $A^4$ & $A^5$ & $A^6$ & $A^7$ & $A^8$ & $A^9$ &  $aAcc$ & $gAcc$\\ \hline
$f_1$ & 85 & - & - & - & - & - & - & - & - & 85.00 & 81.46 \\
$f_2$ & 85 & 0 & - & - & - & - & - & - & - & 78.46 & 66.29 \\
$f_3$ & 85 & 0 & 0 & - & - & - & - & - & - & 72.86 & 57.15 \\
$f_4$ & 85 & 0 & 0 & 0 & - & - & - & - & - & 68.00 & 50.61 \\
$f_5$ & 85 & 0 & 0 & 0 & 0 & - & - & - & - & 63.75 & 45.58 \\
$f_6$ & 85 & 0 & 0 & 0 & 0 & 0 & - & - & - & 60.00 & 41.55 \\
$f_7$ & 85 & 0 & 0 & 0 & 0 & 0 & 0 & - & - & 56.67 & 38.22 \\
$f_8$ & 85 & 0 & 0 & 0 & 0 & 0 & 0 & 0 & - & 53.68 &  35.42 \\
$f_9$ & 85 & 0 & 0 & 0 & 0 & 0 & 0 & 0 & 0 & 51.00 &  33.01 \\\hline
Avg &  &  &  &  &  &  &  & & & 65.49 &  49.92\\
\toprule
\end{tabular}
\end{table*}
\begin{table*}[t]
\setlength{\tabcolsep}{1mm}
\centering
\scriptsize
\caption{Detailed accuracies setting of the corner case \textbf{Greedy}
In this case, the model only focuses on the novel task and does not care the forgetting of the old task.
$f_i$: model after trained on task $\mathcal{T}_i$.
$A^i$: accuracy on task $\mathcal{T}_i$ of the corresponding model $f$.}
\label{table:greedy}
\begin{tabular}{c|ccccccccccc}
\midrule
(\%)& $A^1$ & $A^2$ & $A^3$ & $A^4$ & $A^5$ & $A^6$ & $A^7$ & $A^8$ & $A^9$ &  $aAcc$ & $gAcc$\\ \hline
$f_1$ &85 & - & - & - & - & - & - & - & - & 85.00 & 81.46 \\
$f_2$ &0 & 100 & - & - & - & - & - & - & - & 7.69 &  22.01 \\
$f_3$ &0 & 0 & 100 & - & - & - & - & - & - & 7.14 & 16.38 \\
$f_4$ &0 & 0 & 0 & 100 & - & - & - & - & - & 6.67 & 13.49 \\
$f_5$ &0 & 0 & 0 & 0 & 100 & - & - & - & - & 6.25 & 11.59 \\
$f_6$ &0 & 0 & 0 & 0 & 0 & 100 & - & - & - & 5.88 & 10.23 \\\
$f_7$ &0 & 0 & 0 & 0 & 0 & 0 & 100 & - & - & 5.55 & 9.17 \\
$f_8$ &0 & 0 & 0 & 0 & 0 & 0 & 0 & 100 & - & 5.26 &  8.33 \\
$f_9$ &0 & 0 & 0 & 0 & 0 & 0 & 0 & 0 & 100 & 5.00 & 7.64 \\\hline
Avg &  &  &  &  &  &  &  & & & 14.94 &  20.03 \\
\toprule
\end{tabular}
\end{table*}
\begin{table}[t]
\centering
\scriptsize
\setlength{\tabcolsep}{1mm}
\caption{Detailed accuracies setting of the corner case \textbf{Greedy-NF}.
On top of the `Greedy', we keep the novel tasks' performance (\textbf{N}on-\textbf{F}orgetting).
$f_i$: model after trained on task $\mathcal{T}_i$.
$A^i$: accuracy on task $\mathcal{T}_i$ of the corresponding model $f$.}
\label{table:greedy_nf}
\begin{tabular}{c|ccccccccccc}
\midrule
(\%)& $A^1$ & $A^2$ & $A^3$ & $A^4$ & $A^5$ & $A^6$ & $A^7$ & $A^8$ & $A^9$ &  $aAcc$ & $gAcc$\\ \hline
$f_1$ &85 & - & - & - & - & - & - & - &-  & 85.00 & 81.46    \\
$f_2$ &0 & 100 & - & - & - & - & - & - & - & 7.69 & 22.01   \\
$f_3$ &0 & 100 & 100 & - & - & - & - & - & - & 14.29 & 32.76   \\
$f_4$ &0 & 100 & 100 & 100 & - & - & - & - & - & 20.00 & 40.46   \\
$f_5$ &0 & 100 & 100 & 100 & 100 & - & - & - & - & 25.00  & 46.37  \\
$f_6$ &0 & 100 & 100 & 100 & 100 & 100 & - & - & - & 29.41 & 51.12   \\
$f_7$ &0 & 100 & 100 & 100 & 100 & 100 & 100 & - &  - & 33.33 & 55.03  \\
$f_8$ &0 & 100 & 100 & 100 & 100 & 100 & 100 & 100 & - & 36.84 & 58.33   \\
$f_9$ &0 & 100 & 100 & 100 & 100 & 100 & 100 & 100 & 100 & 40.00 & 61.16  \\\hline
avg & &  &  &  &  &  &  &  & & 32.40 & 49.86 \\
\toprule
\end{tabular}
\end{table}

\section{Detailed Network Structures}
\subsection{Backbone}
In this work, we conduct experiments on the vanilla Vision Transformer (ViT)~\cite{vit}.
To compare with the existing FSCIL methods fairly, we modified the ViT configuration to match the number of parameters.
Specifically, we adjust the feature dimension and number of heads of the ViT backbone.
See \Cref{table:backbone} for the detailed configuration.

\subsection{The MLPs used in this work}
For the feature rectification (FR) module proposed in the main paper, we use MLPs to construct the FR module.
We use the MLP structure in the Vision Transformer, \ie for an input $x$, the MLP process it like this: $FC_1\rightarrow \text{GELU} \rightarrow LN \rightarrow FC_2$,
where $FC_1$ and $FC_2$ are two fully connected layers with the same dimension for both input and output.
GELU is the activated function and $LN$ is the LayerNorm normalization.
FR modules are lightweight and only consist of 0.6M parameters for all three datasets.
\section{Dataset statistics}
In this work, we use three datasets to evaluate our proposed method.

\noindent\textbf{CIFAR-100~\cite{cifar100}.}
There are 50,000 images for training and 10,000 images for testing provided in the CIFAR-100 dataset.
There exist 100 classes in total.
For FSCIL, we follow pioneer works to select the first 60 classes as the base task for training.
The rest 40 classes are split into 8 tasks and each task employs a 5-way 5-shot learning.
The resolution of CIFAR-100 dataset is 32$\times$32.

\noindent\textbf{\emph{mini}ImageNet~\cite{mini}.}
With 100 classes, the \emph{mini}ImageNet dataset has a total of 50,000 images for training and 10,000 images for testing.
The resolution of \emph{mini}ImageNet dataset is 84$\times$84.
For FSCIL, we follow prior works to select the first 60 classes for the base training and the rest 40 classes will be separated into 8 tasks.

\noindent\textbf{CUB200~\cite{cub200}.}
In this work, we use `CUB-200' to refer to the CUB-200-2011 dataset, which contains 11,788 images of 200 different types of birds.
The official training set contains 5994 images for training and there are 5794 samples for testing.
The resolution of CUB-200 dataset is 224$\times$224.
Following previous works, we use the first 100 classes to form the base task, and the rest 100 classes are used to form 10 novel classes.
Each novel task follows a 10-way 5-shot setting for FSCIL.

\begin{table}[t]
\centering
  \scriptsize
  \tabcolsep=0.15cm
  \caption{Details of the adjusted backbone. We slightly adjust the vanilla ViT to match the parameters of the ResNet-18.}
  \begin{tabular}{c|cccc}
  \bottomrule[1pt]
  Model     & \multicolumn{1}{c}{\begin{tabular}[c]{@{}c@{}}Embed\\ Dim\end{tabular}} & \multicolumn{1}{c}{\#Heads} & \multicolumn{1}{c}{\#Layers} & \multicolumn{1}{c}{\#Params}  \\ \hline
  Vit-Tiny\cite{deit}  & 192 & 3       & 12       & 5M       \\
  Vit-Small\cite{deit} & 384 & 6       & 12       & 22M       \\
  Vit-Base\cite{vit,deit}  & 768  & 12      & 12       & 86M \\ \hline
  ResNet-18  & 512  & -       & -        & 11M      \\
  Ours      & 256   & 8       & 12       & 10M \\
  \bottomrule[1pt]                                                           
  \end{tabular}
  \label{table:backbone}
  \end{table}

\begin{figure*}[h]
    \centering
      \includegraphics[width=\linewidth]{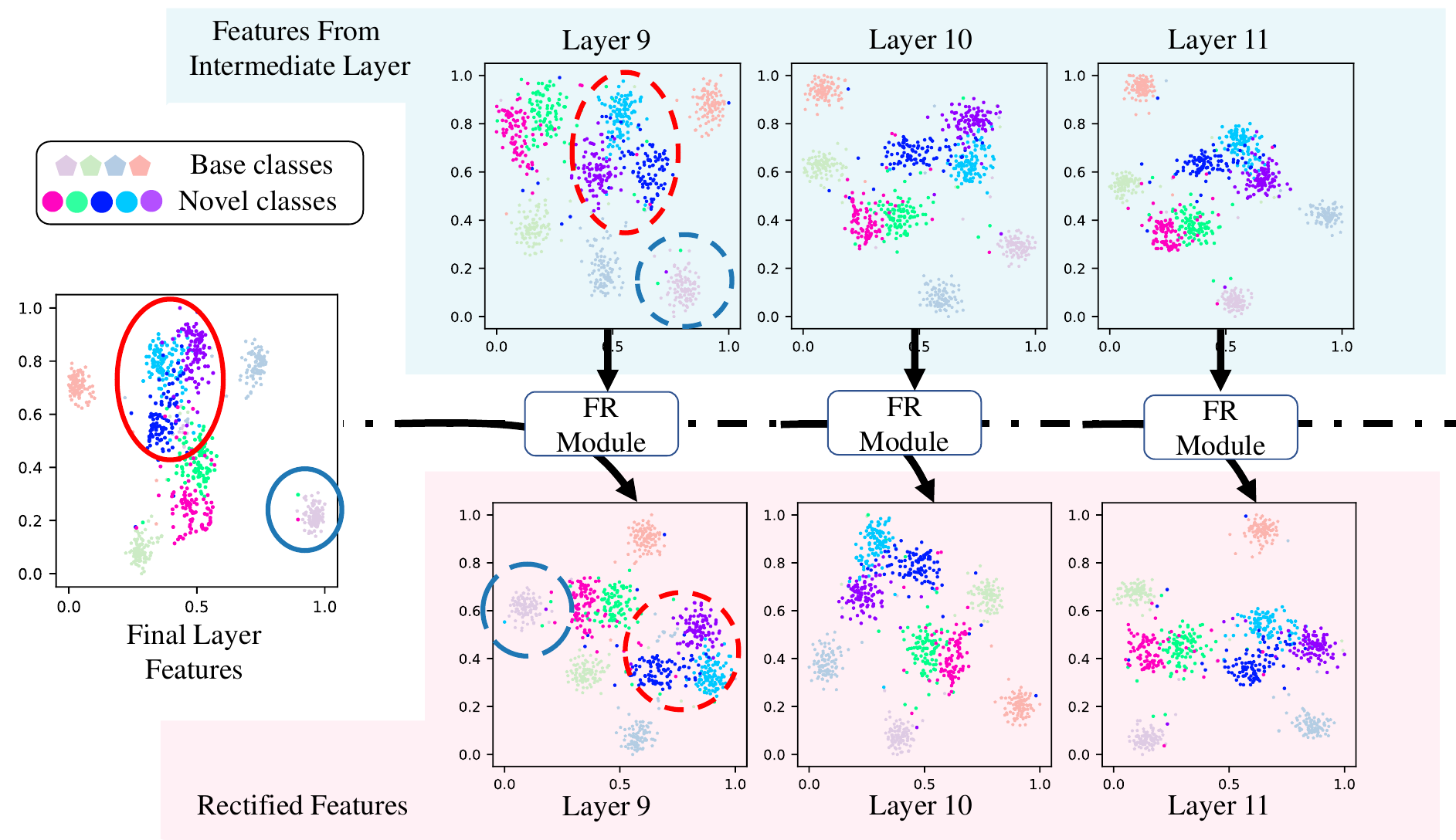}
      \caption{
      Visualization results of the intermediate features, final features and rectified features on the test set of \emph{mini}ImageNet using t-SNE\cite{tsne}.
      This figure is an extension of the Figure \textcolor{red}{3} in the main paper.
      It is shown in this figure that the final features of novel classes are less discriminative compared with the intermediate features (\textcolor{red}{red} circles).
      However, Since the shallow features are more scattered, the performance of base classes will be worse (\textcolor{myBlue}{blue} circles).
      We also show the results of FR modules (\ie rectified features).
      It is clear that the rectified features absorb the advantages of both and perform better.
      Best viewed in color.
      }
    \label{fig:exp_supp_vis_tsne}
  \end{figure*}

\section{Visualization}
Corresponding to Figure \textcolor{red}{3} in the main paper, we further show the t-SNE~\cite{tsne} visualization of the rectified features as well as the original raw features from the Vit backbone.
The visualization results are shown in \cref{fig:exp_supp_vis_tsne}.
As discussed in the main paper, we see that the features from deep layers are more aggregated for the base classes (light-colored scatters in the figure), but their discriminating ability for novel classes is poor; on the contrary, although the shallow features are relatively more dispersed, causing the poor discriminating ability for base classes, the dispersion makes it more discriminative for novel classes.
The FR module is proposed to take advantage of both features, \ie being more discriminative on novel classes while keeping the base-class performance.
The rectified features are shown in the lower part.
As shown in the figure, the rectified features are clearly more distinguishable for different novel classes (red circle in the figure).
At the same time, the proposed training strategy effectively constrains base features and the rectified features on base class are clearly more aggregated than that of the raw intermediate features (blue circle in the figure).
A similar conclusion can be made across different layers shown in the figure.

\section{More Experimental Results}
In this section, we provide more experimental results to show the effectiveness of our method.
Firstly, we show the CIFAR-100 experiment results in \cref{sec:CIFAR100}.
Secondly, a detailed study of novel class performance is shown in \cref{sec:novel}.
Later, more ablation studies including losses ablation on CUB-200, and sensitivity analysis on hyper-parameters are shown in \cref{sec:more_ab}.
\begin{table*}[ht] 
    \scriptsize
	\centering
				\caption{Comparison with FSCIL methods on CIFAR-100. 
            The best and second best results are marked \textbf{bolded} and \underline{underlined}.
            Methods with\dag are reported with the original results in their paper. Other results are reimplemented by us.}
			\begin{tabular}{lccccccccccc}
				\toprule
				\multicolumn{1}{l}{\multirow{2}{*}{\bf Methods}} & \multicolumn{9}{c}{\bf $aAcc$ in each task (\%)} &  &  \\ 
				\cmidrule{2-10}
				& \bf 1      & \bf 2      & \bf 3    & \bf 4     & \bf 5  & \bf 6     & \bf 7      &\bf 8  & \bf 9    & \bf $aAcc$ & $gAcc$ \\ 
				\midrule
				TOPIC~\cite{topic}\dag  & 64.1	&55.9	&47.1	&45.2	&40.1	&36.4	&34.0	&31.6	&29.4&	42.62&  - \\
				Self-promoted  &	64.1	&65.9	&61.4	&57.5	&53.7	&50.8	&48.6	&45.7	&43.3	&54.52&  -   \\
                BiDistFSCIL~\cite{BiDistFSCIL}\dag & 79.5 & 75.4 & 71.8 & 68.0 & 65.0 & 62.0 & 60.2 & 57.7 & 55.9 & 66.14 & - \\
                MetaFSCIL~\cite{metafscil}\dag & 74.5 & 70.1 & 66.8 & 62.8 & 59.5 & 56.5 & 54.4 & 52.6 & 50.0 & 60.79 & - \\
                DSN~\cite{dsn}\dag & 73.0 & 68.8 & 64.8 & 62.6 & 59.4 & 57.0 & 54.0 & 51.6 & 50.0 & 60.14 & - \\
				CEC~\cite{cec}  & 73.2 & 69.1 & 65.4 & 61.2 & 58.0 & 55.5 & 53.2 & 51.3 & 49.2 & 59.56 & 50.85  \\
                CLOM~\cite{clom} & 74.2 & 69.8 & 66.1 & 62.2 & 59.0 & 56.1 & 54.0 & 51.9 & 49.8 & 60.35 & 50.36 \\
				C-FSCIL~\cite{cfscil} & 76.8 & 72.0 & 67.2 & 63.1 & 59.7 & 56.7 & 54.4 & 51.9 & 49.7 & 61.28 & 50.62 \\
				DF-Replay~\cite{data_free} & 72.9 & 68.1 & 65.1 & 61.6 & 58.5 & 54.8 & 52.8 & 51.0 & 49.1 & 59.31 & 51.53  \\
                LIMIT~\cite{limit}  & 76.1 & 71.7 & 67.3 & 63.2 & 60.1 & 57.3 & 55.2 & 53.0 & 50.8 & 61.64 & 52.16  \\
                SAVC~\cite{savc} & 78.4 & 72.7 & 68.4 & 64.0 & 61.2 & 58.3 & 56.0 & 54.1 & 51.7 & 62.76 & 53.29  \\
                FACT~\cite{fact} & 78.5 & 72.7 & 68.8 & 64.5 & 61.3 & 58.6 & 56.8 & 54.4 & 52.3 & 63.12 & 54.56  \\ 
                SubNetwork~\cite{subnetwork} & 80.0 & 75.7 & 71.6 & 67.8 & 64.7 & 61.4 & 59.3 & 57.1 & 55.1 & 64.85 & 56.63 \\
                S3C~\cite{s3c} & 78.0 & 73.9 & 70.2 & 66.1 & 63.3 & 60.1 & 58.3 & 56.7 & 54.0 & 64.51 & 58.04 \\
                ALICE~\cite{alice} & 80.3	& 71.9	& 67.0	& 63.2	& 60.7	& 58.3	& 57.3	& 55.5	& 53.7	& 63.10 & 59.72 \\
                NC-FSCIL~\cite{neural_collapse} & 82.9	& 77.5	& 73.7	& 69.0	& 65.6 & 62.0 & 59.7 & 57.8 & \underline{55.4}	& \textbf{67.06} & \underline{59.73} \\
				\midrule
				Ours &  82.9	& 76.3	& 72.9	& 67.8	& 65.2	& 62.0	& 60.7	& 58.8	& \textbf{56.6}	& \underline{67.02} & \textbf{60.32}\\
				\bottomrule
			\end{tabular}
		\label{table:cifar_supp}
\end{table*}

\begin{figure}[h]
    \centering
      \includegraphics[width=0.7\linewidth]{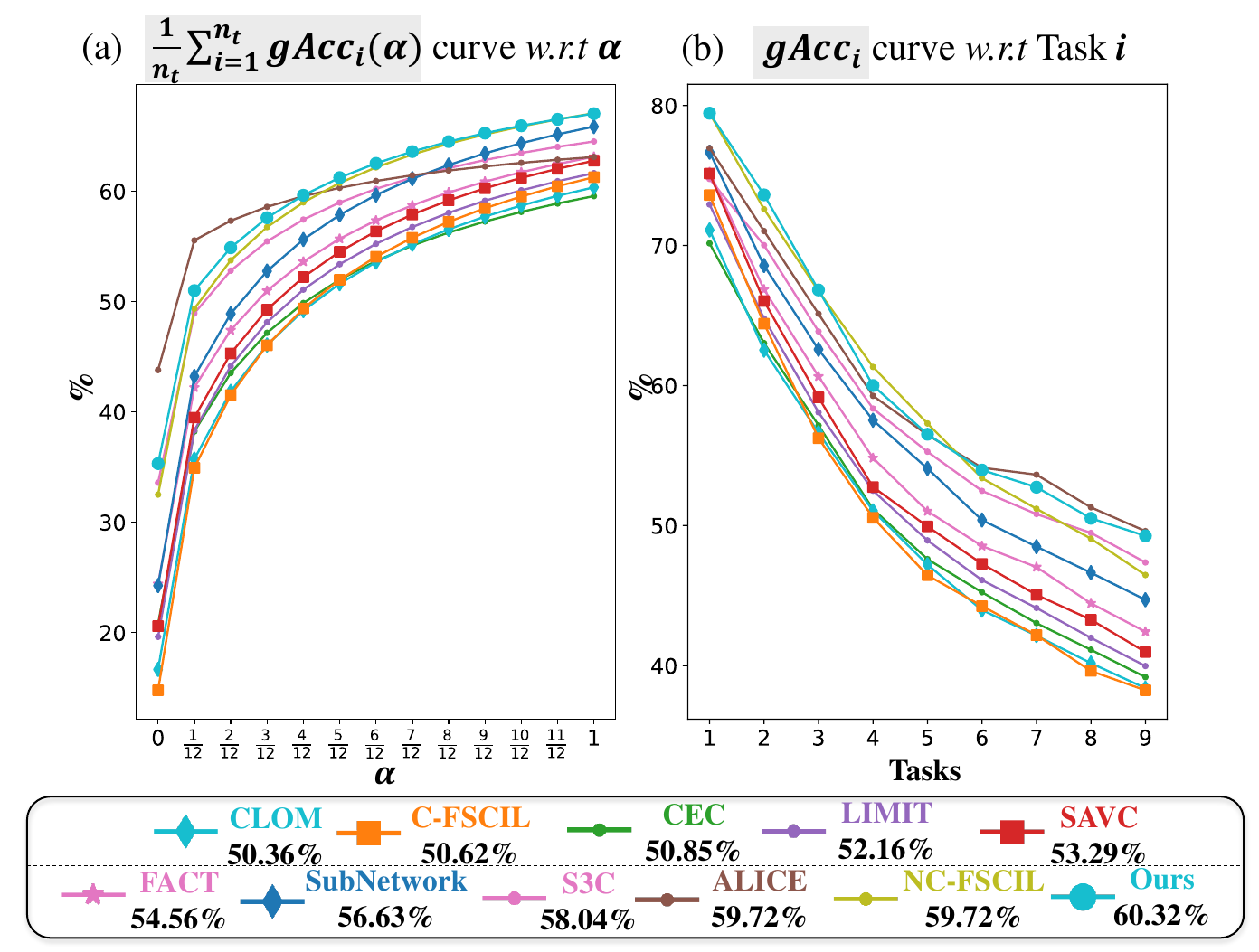}
      \caption{
      FSCIL performances on the CIFAR-100 dataset are shown in \textit{generalized accuracy}.
      In the legend, we show the average AUC across tasks(Eq. (\textcolor{red}{6)} in the main paper) of each method.
      \textbf{(a)}: The $gAcc$ curve (averaged across all $n_t$ tasks) v.s. param $\alpha$.
      \textbf{(b)}: The $gAcc$ AUC of each task $\mathcal{T}_i$ (see Eq. (\textcolor{red}{5)}) in the main paper) at each task.
      Best viewed in color.
      }
    \label{fig:exp_cifar_all}
  \end{figure}

\subsection{More Results on CIFAR-100}
\label{sec:CIFAR100}
Experimental results conducted on the CIFAR-100 dataset are shown in \Cref{table:cifar_supp}.
Although our method does not outperform NC-FSCIL in terms of average acc ($aAcc$), our method achieves higher generalized accuracy ($gAcc$) than NC-FSCIL.
When we look into the detailed accuracy in each task, we can see that as the tasks proceed, our method achieves higher accuracies than NC-FSCIL (task $\mathcal{T}_9$).
This is mainly because our method gets better novel class performance than NC-FSCIL.
The detailed evaluation on $gAcc$ of the CIFAR-100 dataset is shown in \cref{fig:exp_cifar_all}.
We can observe similar phenomena discussed in the main paper.
For example, a recent SOTA SubNetwork~\cite{subnetwork} does not necessarily perform better than S3C~\cite{s3c} when considering the novel class performance.
In fact, it is shown in the figure that although SubNetwork achieves better $aAcc$ ($\alpha=1$) than S3C, the S3C instead achieves a far larger AUC than SubNetwork.
Similar observation can be found when comparing method C-FSCIL~\cite{cfscil} with DF-replay~\cite{data_free}.
It is clear that the $gAcc$ curve and the corresponding AUC can offer a more fine-grained and more detailed perspective when evaluating FSCIL methods.
 
\begin{table}[ht] 
        \scriptsize
		\centering
  \setlength{\tabcolsep}{4pt}
    \caption{Novel class performance in incremental tasks on \emph{mini}ImageNet.}
			\begin{tabular}{lccccccccccc}
				\toprule
				\multicolumn{1}{l}{\multirow{2}{*}{\bf Methods}} & \multicolumn{9}{c}{\bf Novel class performance in incremental tasks (\%)}\\ 
				\cmidrule{2-10}
			       & \bf 2      & \bf 3    & \bf 4     & \bf 5  & \bf 6     & \bf 7      &\bf 8  & \bf 9 &Avg& $aAcc$ & $gAcc$\\ 
				\midrule
   CEC\cite{cec} & 21.4 & 21.3 & 19.9 & 20.0 & 18.0 & 17.0 & 17.5 & 17.7 & 19.1 & 57.91 & 48.64   \\
   Self-Prompted\cite{sppr}&6.6&6.3&5.0&4.1&6.4&6.8&6.7&7.0& 6.11 &52.62&41.53\\
   LIMIT\cite{limit}&31.0&23.8&22.3&21.7&19.4&19.0&19.4&20.0&22.01&58.15&49.47\\
   FACT\cite{fact}&18.8&16.6&15.0&14.1&12.4&11.4&11.7&12.2&14.03&60.79&49.56\\
   CLOM\cite{clom}&6.4&10.5&11.2&11.5&10.6&9.7&10.7&11.6&10.28&57.23&46.11\\
   C-FSCIL\cite{cfscil}&36.2&34.6&35.3&35.8&33.7&31.5&32.2&32.6&33.99&55.39&50.46\\
   S3C\cite{s3c}&25.0&26.0&27.9&25.7&23.7&22.5&23.7&23.5&24.75&62.74&53.70\\
   SAVC\cite{savc} & 31.6 & 27.1 & 26.6 & 26.3  & 22.8  &	21.4 & 21.6 & 22.4& 24.98 & 66.66 & 56.35   \\
   BiDistFSCIL\cite{BiDistFSCIL} & 32.2 & 32.4 & 29.3 & 31.2 & 26.6  & 25.8 & 26.2 & 25.8 & 28.69 & 61.32 & 53.49 \\
   Regularizer\cite{regularizer} & 47.2  & 44.8	& 37.1  & 34.0  & 39.8  & 22.0	& 25.4	& 26.2 & 34.56 & 62.82 & 55.62 \\
   SubNetwork\cite{subnetwork} 	& 35.4 & 33.1 &  30.1 & 29.0 & 27.0 & 26.2 & 26.4 & 27.8 & 29.38 &64.31 & 55.88  \\
   NC-FSCIL\cite{neural_collapse}  & 52.0	& 45.4	& 42.9	& 40.2  & 37.4 & 34.4 & 32.7 & 31.5 & 39.56& 67.90 & 60.93 \\
                \hline
   Vit-Baseline & 22.4  & 24.0 & 24.3 & 23.6  & 21.8 & 21.3 & 22.5 & 22.6& 22.81 & 68.74 & 57.87\\
   +YourSelf (Ours)  & 51.4  & 49.0 & 44.1 & 43.5  & 40.3 & 38.5 &38.1 & 39.0 &\bf 42.99 & \bf 68.80 & \bf 62.20 \\
				\hline
			\end{tabular}

		\label{table:novel_mini}
\end{table}
\begin{table}[pt] 
        \scriptsize
		\centering
  \setlength{\tabcolsep}{4pt}
    \caption{Novel class performance in incremental tasks on CIFAR-100.}
			\begin{tabular}{lccccccccccc}
				\toprule
				\multicolumn{1}{l}{\multirow{2}{*}{\bf Methods}} & \multicolumn{9}{c}{\bf Novel class performance in incremental tasks (\%)}\\ 
				\cmidrule{2-10}
			       & \bf 2      & \bf 3    & \bf 4     & \bf 5  & \bf 6     & \bf 7      &\bf 8  & \bf 9 & Avg & $aAcc$ & $gAcc$ \\ 
				\midrule
   CEC\cite{cec}&30.2&27.3&22.0&21.4&22.1&21.9&21.5&20.9&23.41&59.56&50.85\\
   CLOM\cite{clom}&22.6&22.0&18.6&17.5&16.7&17.7&17.5&17.4&18.75&60.35&50.36\\
   C-FSCIL\cite{cfscil}&23.4&16.3&14.0&13.2&16.3&16.9&15.8&17.2&16.64&61.27&50.62\\
   DF-Replay\cite{data_free}&43.0&33.6&27.9&27.1&22.4&22.4&23.4&22.3&27.76&59.31&51.53\\
   LIMIT\cite{limit}&27.2&24.4&21.3&21.0&21.0&21.2&20.5&20.1&22.09&61.64&52.16\\
   SAVC\cite{savc}&30.0&25.6&20.0&21.7&22.3&22.3&22.3&21.2&23.18&52.76&53.29\\
   FACT\cite{fact}&35.2&31.0&26.6&25.2&25.7&26.8&25.1&24.2&27.48&63.12&54.56\\
   S3C\cite{s3c}&49.2&40.8&36.2&35.0&35.2&35.2&35.5&35.2&37.79&64.50&58.04\\
   NC-FSCIL\cite{neural_collapse}&46.0&41.4&39.1&36.4&34.0&33.6&32.1&30.0&36.58&\bf67.06&59.72\\
                \hline
   Vit-Baseline &23.6&16.5&15.1&13.5&14.0&14.3&14.5&15.0&15.81&66.35&54.21\\
   +YourSelf (Ours) & 58.8&44.7&37.3&34.8&35.8&32.3&34.5 &35.8&\bf39.25&67.02&\bf60.32\\
				\hline
			\end{tabular}

		\label{table:novel_cifar}
\end{table}
\subsection{Novel classes performance}
\label{sec:novel}
As a supplement to Sec. \textcolor{red}{5.2} in the main paper, we also provide results of only novel classes ($gAcc(0)$) learning only across incremental learning tasks.
Results on \emph{mini}ImageNet and CIFAR-100 are shown in \cref{table:novel_mini} and \cref{table:novel_cifar} respectively.
Similar to the conclusion stated in Sec. \textcolor{red}{5.2}, we can see from the tables that the $aAcc$ metric can not effectively reflect the change of novel class performance.
For example, in \cref{table:novel_mini}, models CLOM and C-FSCIL share similar $aAcc$ performance while their novel class performance is not even close. 
Similar conclusions can also be summarized by comparing FACT and S3C in \cref{table:novel_cifar}.
This validates the necessity of our proposed $gAcc$, which can effectively show the performance gap on novel tasks, complementary to the existing $aAcc$ metric. 
Experimental results in the two tables also show the effectiveness of our proposed method. 
It is clear that our proposed method effectively boosts the novel class performance compared to the baseline in both datasets.
Furthermore, our proposed Yourself method outperforms other methods on novel tasks on average. 

\subsection{More Ablation Studies}
\label{sec:more_ab}
\noindent\textbf{Ablation studies on CUB-200.}
The ablation study regarding the effectiveness of each loss on CUB-200 is shown in \Cref{table:ablate_cub}.
We can see that the introduction of the FR module as well as the loss $\mathcal{L}_{cos}$ and loss $\mathcal{L}_{NovCE}$  can bring 1.26\% improvement on $gAcc$.
Additionally, the $gAcc$ performance can be boosted by 0.4\% and 1.47\% when introducing the loss $\mathcal{L}_{CR}$ and loss $\mathcal{L}_{IR}$ respectively.
This indicates the effectiveness of each loss we proposed.

\begin{table*}[ht]
  \centering
  \scriptsize
  \caption{Ablation studies on CUB-200 dataset.}
    \begin{tabular}{ccccccccccccccccc}
    \toprule
    \multirow{2}*{{FR}} & \multirow{2}*{$\mathcal{L}_{Cos}$ +$\mathcal{L}_{NovCE}$} & \multirow{2}*{{$\mathcal{L}_{CR}$}} & \multirow{2}*{\textbf{$\mathcal{L}_{IR}$}} & \multicolumn{11}{c}{$aAcc$ in each task (\%)} & \multirow{2}{*}{$aAcc$} & \multirow{2}{*}{$gAcc$}  \\
\cline{5-15}  &       &       &  &  1 & 2 & 3 & 4 & 5 &  6 & 7 & 8 & 9 & 10 & 11 \\
    \hline
          &       &   & & 84.0 & 79.6 & 76.4 & 72.5 & 70.1 & 67.4 & 66.3 & 65.2 & 63.5 & 63.2 & 52.3 & 70.16 & 62.37  \\
    \checkmark     & \checkmark    &  & & 83.4 & 79.9 & 76.8 & 73.2 & 70.8 & 68.0 & 66.9 & 66.3 & 64.3 & 64.2 & 63.3 & 70.57 & 63.63  \\
    \checkmark     & \checkmark        & \checkmark     &   & 83.4 & 78.7 & 75.7  & 72.1  & 70.2  & 67.3 & 65.7  & 65.3  & 63.4 & 63.6 & 62.7 & 69.93 & 64.03 \\
    \checkmark     & \checkmark         & \checkmark     &\checkmark    & 83.4 & 77.0 & 75.3 & 72.2 & 69.0 & 66.8 & 66.0 & 65.6 & 64.1 & 64.5 & 63.6 & 69.85 & 65.50  \\
    \bottomrule
    \end{tabular}
  \label{table:ablate_cub}
\end{table*}

\begin{figure}[t]
    \centering
      \includegraphics[width=\linewidth]{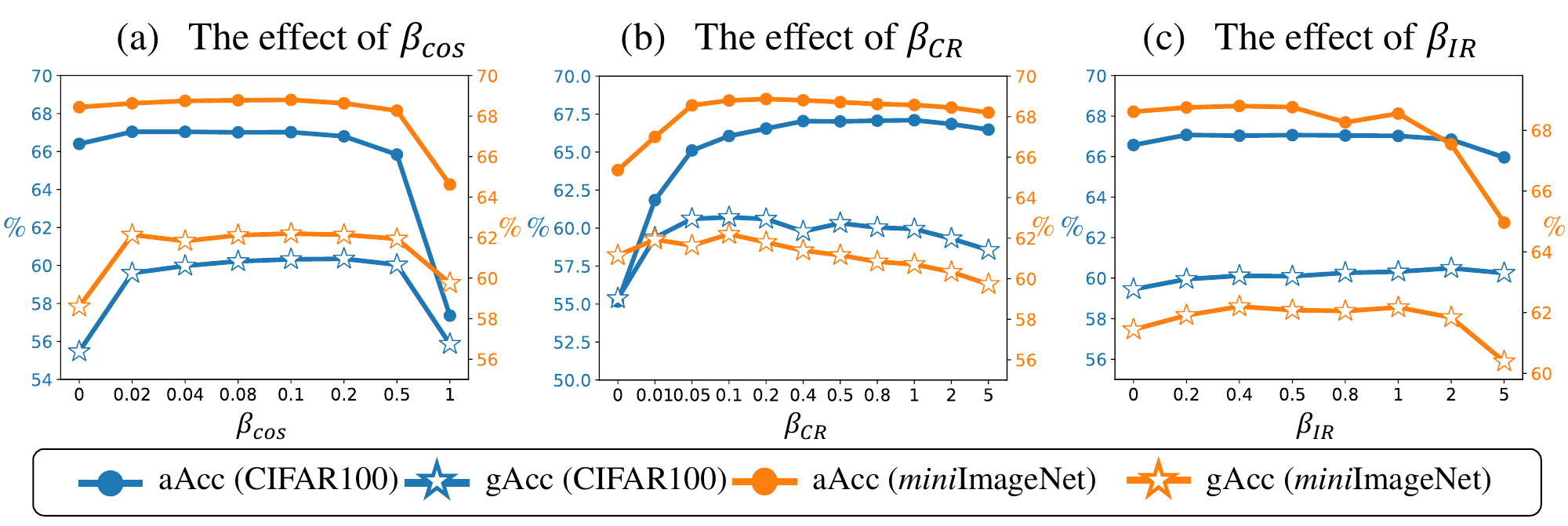}
      \caption{
      Ablation studies on the impact of three hyper parameters $\beta_{cos}$, $\beta_{CR}$ and $\beta_{IR}$.
      We use \textcolor{myOrange}{orange} to indicate the results on \emph{mini}ImageNet and \textcolor{myBlue}{Blue} to indicate the results from CIFAR-100.
      In each sub-figure, the y-axis for the miniImageNet dataset is \textcolor{myOrange}{on the right in orange}, while the y-axis for the CIFAR-100 dataset is \textcolor{myBlue} {on the left in blue}.
      We report both $aAcc$ ($\bullet$) and $gAcc$ (\hollowstar) results in the figure.
      Best viewed in color.
      }
    \label{fig:exp_supp_sensitivity}
  \end{figure}

\noindent\textbf{The impact of trade-off parameters.}
Recall that the overall loss function described in the main paper is as follows.
\begin{equation}
    \mathcal{L} = \beta_{cos}(\mathcal{L}_{cos} + \mathcal{L}_{NovCe}) + \beta_{CR}\mathcal{L}_{CR} + \beta_{IR}\mathcal{L}_{IR}
\end{equation}
In order to study the effect of different parameter values on the final performance, we conduct experiments on three hyper-parameters $\beta_{cos}$, $\beta_{CR}$ and $\beta_{IR}$.
Experiments are conducted on CIFAR-100 and \emph{mini}Imagenet.
Results are shown in \cref{fig:exp_supp_sensitivity}.
\begin{itemize}

\item\textbf{The effect of $\beta_{cos}$.} We can see in the \cref{fig:exp_supp_sensitivity} (a) that the $\mathcal{L}_{cos}$ and  $\mathcal{L}_{NovCe}$ can effectively improve the $gAcc$ without altering the $aAcc$ significantly.
When we adjust the weight $\beta_{cos}$ ranging from 0 to 1, we can see the $aAcc$ remains almost unchanged (except for the outlier of $\beta_{cos} = 1$ on \emph{mini}ImageNet) but the $gAcc$ first rise and then fall.
By default, we set the $\beta_{cos} = 0.1$ in all other experiments in this work.

\item\textbf{The effect of $\beta_{CR}$.}
We conduct experiments by setting the hyper-parameter $\beta_{CR}$ of the loss $\mathcal{L}_{CR}$ ranging in $[0,1]$.
From \cref{fig:exp_supp_sensitivity} (b), it can be concluded that the introduction of the loss $\mathcal{L}_{CR}$ can bring up to 4\% performance gain on $gAcc$.
Furthermore, we can see that for both $aAcc$ and $gAcc$, the performance rises first and goes down when the $\beta_{CR}$ gets larger.
We set $\beta_{CR} = 0.1$ as our default setting on \emph{mini}ImageNet and $\beta_{CR} = 0.5$ on CIFAR-100,  which is applied to all experiments in this work.
 
\item\textbf{The effect of $\beta_{IR}$.}
Similarly, we show the experiment about the effect of different $\beta_{IR}$.
When changing values of $\beta_{IR}$ form 0 to 1, the $gAcc$ is rising up and then falling down while the $aAcc$ on \emph{mini}ImageNet is falling down.
It shows that with proper weights, the $\mathcal{L}_{IR}$ can effectively boost the $gAcc$ without doing much harm to the $aAcc$.
We set $\beta_{IR} = 1$ as the default setting for CIFAR-100 and $\beta_{IR} = 0.5$ for \emph{mini}ImageNet.
 
\end{itemize}

\section{Discussion about our $gAcc$ and other alternative metrics in FSCIL}
\noindent\textbf{Harmonic Accuracy~\cite{alice}.}
It is worth mentioning that in previous work Alice~\cite{alice} proposed a metric called \textit{harmonic accuracy} based on the harmonic mean to tackle the problem of the base-novel balance problem.
The \textit{harmonic accuracy (hAcc)} at task $\mathcal{T}_i$ can be denoted as:
\begin{equation}
    hAcc_i = \frac{2A^1_i A^{novel}_i}{A^1_i + A^{novel}_i},
    \label{eq:hacc}
\end{equation}
where $A^1_i$ is the base class accuracy and the $A^{novel}_i = \frac{1}{i-1} \sum_{j=2}^i A_i^j$ is the averaged performance of the novel tasks.
The $hAcc$ is derived from the harmonic mean which was originally designed for calculating the average of proportions (\eg velocity).
Direct use of the harmonic mean to calculate accuracy loses its clear physical meaning.
Further, the $hAcc$ has the following limitation when evaluating the FSCIL methods.
\begin{itemize}
\item \textbf{Symmetry Issue.}
    The harmonic mean resembles the arithmetic mean, where in the FSCIL context, the arithmetic mean represents the `task-wise accuracy' (as detailed in Sec. \textcolor{red}{3.1} of the main paper). 
    Both arithmetic mean and harmonic mean treat the performance from base tasks and novel tasks equally.
    As shown in \cref{eq:hacc}, the roles of $A^1_i$ and $A^{novel}_i$ are symmetry.
    Unlike our $gAcc$, this will ignore the difficulty of the base task and over-emphasize the performance of the novel classes.
    A better solution to balancing this `base-novel' dilemma is to analyze the performance from more fine-grain perspectives like our proposed $gAcc$.
    \item \textbf{Heterogenicity.}
    When comparing different methods using a metric, it's routine to utilize the absolute difference between their values to signify the performance gap. However, when employing $hAcc$ for the partial derivative of $A^{novel}_i$, the equation
    \begin{equation}
    \frac{\partial (hAcc)}{\partial A^{novel}_i} = \frac{2 (A^1_i)^2}{(A^1_i + A^{novel}_i)^2},
    \end{equation}
    indicates a non-linear change rate in $hAcc$.
    This non-linearity is prominent when $(A^1_i + A^{novel}_i) < 1$ (common in FSCIL), causing the derivative to exceed one. Essentially, with a small $(A^1_i + A^{novel}_i)$ and a constant $A^1_i$ while $A^{novel}_i$ decreases, the reduction in $hAcc$ amplifies. Consequently, when evaluating methods with low $A^{novel}_i$, the absolute difference in $hAcc$ becomes inaccurate due to this amplification.
    Termed as `heterogeneity', this phenomenon hampers the visualization of differences between methods. In contrast, our proposed $gAcc$ exhibits linear changes, ensuring that the absolute difference accurately reflects performance gaps among methods.
    \item \textbf{Instability.}
    It is clear from \cref{eq:hacc} that if $A^{novel}_i$ or $A^1_i$ equals zero, the overall accuracy will be zero (or not defined when we consider the original definition of the harmonic mean).
    This makes the analysis of corner cases (like we do in the main paper and  \cref{sec:corner_cases}) more difficult.

\end{itemize}

\noindent\textbf{Forgetting metrics.}
In addition to metrics like PD and KR stated in Sec. \textcolor{red}{3.3} of the main paper, we further discuss two more forgetting metrics Performance Dropping (PD) and Relative Performance Dropping~\cite{graph}(RPD) here.
Both PD and RPD are metrics to measure the forgetting of the incremental learning methods.
For a particular task $\mathcal{T}_i$, PD evaluates the performance by PD $= A^i_{n_t} - A^i_i$, which calculates the performance drops between the first seen task $\mathcal{T}_i$ and the last task $\mathcal{T}_{n_t}$.
Another similar metric forgetting~\cite{forgetting_metric} also reflects the performance drops but it focuses on strictly the \textit{maximum} drop: $F = \max_j A^i_j -  \min_j A^i_j$.
RPD~\cite{graph} is the extension of PD that calculates the ratio of the PD by the accuracy of the first task, \ie RPD$=\frac{\text{PD}}{A^i_i }$

These metrics show the forgetting of FSCIL methods serve a different unique role in the evaluation and
our proposed $gAcc$ is complementary to them.
Similar to the conclusion in Sec. \textcolor{red}{3.3}, all these forgetting metrics will fail to correctly reflect the poor performance of cases like `Lazy' while ours can effectively capture changes on the novel class performance.

\end{document}